\pdfoutput=1
\documentclass[11pt]{article}

\usepackage[final]{acl}
\usepackage{graphicx}
\usepackage{times}
\usepackage{latexsym}
\usepackage[T1]{fontenc}
\usepackage[utf8]{inputenc}
\usepackage{booktabs}
\usepackage{bm}
\usepackage{multirow}
\usepackage{xurl}
\usepackage{hyperref}
\usepackage{colortbl}
\usepackage{amsmath,amssymb,amsfonts,amsthm}
\usepackage{cases}
\usepackage{subcaption}
\usepackage{makecell}
\usepackage{afterpage}
\makeatletter
\newcommand{\tabfnmark}[1]{\footnotemark\expandafter\xdef\csname tabfn@n@#1\endcsname{\the\value{footnote}}\expandafter\global\expandafter\let\csname tabfn@h@#1\endcsname\Hy@footnote@currentHref}
\newcommand{\tabfntext}[2]{\begingroup\edef\tabfn@save{\the\value{footnote}}\setcounter{footnote}{\csname tabfn@n@#1\endcsname}\expandafter\global\expandafter\let\expandafter\Hy@footnote@currentHref\csname tabfn@h@#1\endcsname\footnotetext{#2}\setcounter{footnote}{\tabfn@save}\endgroup}
\makeatother

\definecolor{okabeblue}{HTML}{0072B2}
\definecolor{okabeorange}{HTML}{E69F00}
\definecolor{okabegreen}{HTML}{009E73}
\definecolor{okabered}{HTML}{CC3311}

\newcommand{\gdiff}[1]{\textcolor{okabegreen!80!black}{\fontsize{8pt}{8pt}\selectfont{$\uparrow$#1}}}
\newcommand{\rdiff}[1]{\textcolor{okabered!80!black}{\fontsize{8pt}{8pt}\selectfont{$\blacktriangle$#1}}}

\newcommand{\secondbest}[1]{#1\rlap{$^{\dagger}$}}

\usepackage{xcolor}

\title{See Every Block Once: Variance-Reduced Post-Training\\
for Masked Diffusion Language Models}

\author{
   Yunzhen He$^{1}$ \qquad
   Hiroaki Yamagiwa$^{2}$\thanks{Work done while at Kyoto University.} \qquad
   Hidetoshi Shimodaira$^{1,3}$ \\
  ${}^1\,$Kyoto University \qquad
  ${}^2\,$SB Intuitions \qquad
  ${}^3\,$RIKEN AIP\\
  \texttt{he.yunzhen.25d@st.kyoto-u.ac.jp,}\\
  \texttt{hiroaki.yamagiwa@sbintuitions.co.jp, shimo@i.kyoto-u.ac.jp}\\
}

\begin{document}
\maketitle

\begin{abstract}
Policy gradient methods are widely used to align large language model outputs with human preferences, but the log-likelihoods they require cannot be directly computed in Masked Diffusion Language Models (MDLMs).
MDLM post-training therefore relies on a surrogate objective based on the Evidence Lower Bound (ELBO), and recent work estimates a blockwise variant of this objective via Monte Carlo methods to support blockwise generation.
However, existing estimation schemes randomly sample the block index, so they may evaluate the same block more than once and leave some blocks unevaluated. This introduces variance arising from block selection into the estimates.
In this work, we propose a full-block estimator that evaluates every block exactly once and averages the results.
We theoretically show that the proposed estimator remains unbiased while eliminating the between-group variance arising from block selection, thereby achieving lower variance than the baseline for the same sample size.
In experiments on five benchmarks spanning mathematics, planning, and coding, our method achieves the highest accuracy on four benchmarks and outperforms the random block sampling baseline by an average of 12.79 points.\footnote{Code is available at \url{https://github.com/githubhyz/See_Every_Block_Once}.}
\end{abstract}

\begin{figure}[t!]
    \centering
    \includegraphics[width=\linewidth]{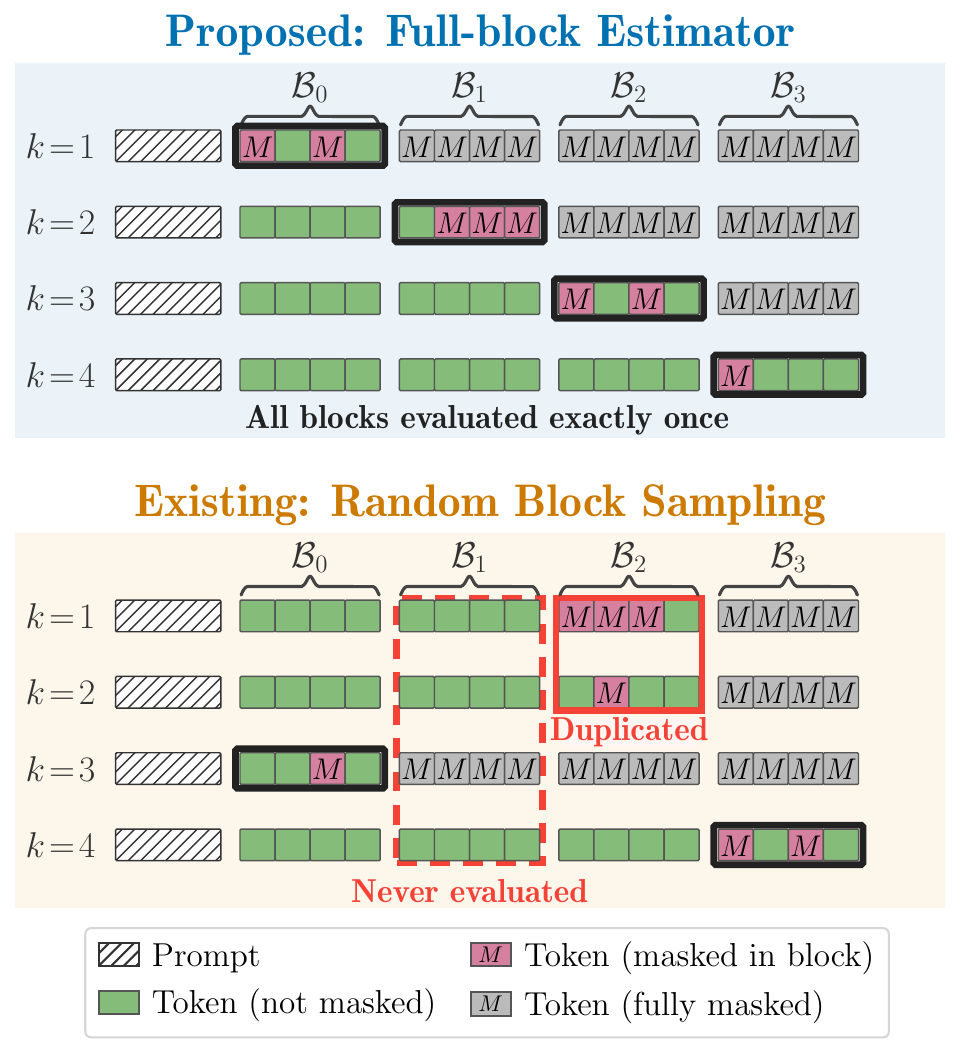}
\caption{
Comparison of block selection in Monte Carlo estimation of the blockwise surrogate objective.
For each Monte Carlo sample index $k$, a block $b$ is sampled, and the corresponding set of token positions is denoted by $\mathcal{B}_b$.
For each block $b$, the masking ratio $t_b$ and the masked positions $\mathcal{M}_{t_b}$ are sampled to compute the blockwise surrogate objective in (\ref{eq:block_obj_intro}).
\textbf{Top:} In our method (full-block evaluation), each sample $k$ corresponds to a distinct block, so that every block is evaluated exactly once. \textbf{Bottom:} In the existing method (random block sampling), each sample independently selects a block, which may cause the same block to be evaluated multiple times ($\mathcal{B}_2$ is selected for both $k{=}1,2$) or some blocks to remain unevaluated ($\mathcal{B}_1$).}
\label{fig:fig1}
\end{figure}

\section{Introduction}

Policy gradient methods are widely used to align large language model (LLM) outputs with human preferences~\cite{NEURIPS2022_b1efde53, NEURIPS2023_a85b405e, shao2024deepseekmath}.
These methods require the log-likelihood of the response, which in auto-regressive language models (ARLMs) can be computed as the sum of conditional log-probabilities over individual tokens.

In contrast, Masked Diffusion Language Models (MDLMs), which have recently gained attention, cannot directly compute log-likelihoods.
MDLMs generate sequences by randomly masking tokens in a sequence and then progressively unmasking them.
The generation process in MDLMs does not uniquely determine the unmasking order, so computing the log-likelihood requires marginalization over all possible orders~\cite{pmlr-v32-uria14}.
For long sequences, this computational cost is prohibitive.

Consequently, MDLM training cannot directly maximize the log-likelihood of the full response and instead optimizes the Evidence Lower Bound (ELBO), as a surrogate objective~\cite{NEURIPS2024_eb0b13cc}.
Recent MDLMs also adopt blockwise generation, which partitions sequences into blocks and generates them left to right~\cite{ICLR2025_7ede97c3, NEURIPS2025_48b383b2}; a corresponding blockwise surrogate objective has also been proposed~\cite{ICLR2026_2a2de7a4, sun2025blockwise}.

This work focuses on the Monte Carlo estimation of the blockwise surrogate objective.
The blockwise surrogate objective is defined as an expectation over the block index $b$, the masking ratio $t_b$ within the block, and the subset of masked positions $\mathcal{M}_{t_b}$:
\begin{align}
    \mathcal{L}_{\text{Block}}
    = \mathbb{E}_{b,\, t_b,\, \mathcal{M}_{t_b}}
    \!\left[
    \ell_{b,t_b,\mathcal{M}_{t_b}}
    \right]
    \label{eq:block_obj_intro}
\end{align}
(see Section~\ref{subsec:block_generation} for the definition).
Here, $\ell_{b,t_b,\mathcal{M}_{t_b}}$ is the token prediction loss at the masked positions within block $b$.
In practice, $\mathcal{L}_{\text{Block}}$ is approximated via Monte Carlo estimation by sampling $b$, $t_b$, and $\mathcal{M}_{t_b}$ a small number of times, and the resulting estimation variance can destabilize policy updates.
In particular, the existing estimation scheme~\citep{ICLR2026_2a2de7a4} that randomly samples the block index $b$ (random block sampling; Fig.~\ref{fig:fig1}, bottom) may select the same block multiple times or leave some blocks entirely unevaluated.

Unlike prior work that focuses on designing policy optimization rules~\cite{NEURIPS2025_52190a03, ICLR2026_f381114c}, improving the surrogate objective~\cite{ICLR2026_2a2de7a4}, or controlling the bias and variance of likelihood approximation~\cite{zhu-etal-2026-llada}, this work examines the design of the Monte Carlo estimator itself while keeping the surrogate objective fixed.
Specifically, instead of randomly sampling the block index, we propose a full-block estimator that evaluates every block exactly once and averages the results (Fig.~\ref{fig:fig1}, top).
Since this approach replaces only the estimator without modifying the blockwise surrogate objective itself, it can be readily integrated into existing MDLM post-training pipelines.

We evaluate our method on five benchmarks (GSM8K, MATH-500, Countdown, Sudoku, and HumanEval) and find that it outperforms baselines on most benchmarks.
Furthermore, compared with random block sampling, our method exhibits higher Spearman rank correlation and lower estimation error relative to reference values treated as ground truth, showing that it suppresses the variance arising from block selection at the estimator level.
These results indicate that the design of the estimator for the surrogate objective is also a viable avenue for improvement in MDLM post-training.

\section{Background}
This section first describes Masked Diffusion Language Models (MDLMs), which form the basis of our method, then formulates blockwise generation used in MDLMs, and finally explains the instability of block selection.

\subsection{Masked Diffusion Language Models}
\label{subsec:mdlm_overview}
Masked Diffusion Language Models (MDLMs) apply an absorbing-state diffusion process to discrete text sequences and generate sequences through mask prediction~\citep{NEURIPS2024_eb0b13cc,NEURIPS2025_48b383b2}.
In the forward process, each token is independently masked according to a diffusion timestep $t \in (0,1]$, and in the reverse process the sequence is recovered by predicting the tokens at masked positions (see Appendix~\ref{app:mdlm_details} for details).

\paragraph{Notation.}
Let $\bm{o}$ denote the response generated by the model for an input prompt $\bm{c}$.
Each response $\bm{o} = (o^{1}, \dots, o^{L})$ consists of $L$ tokens over a vocabulary $\mathcal{V}$.
We denote by $\bm{o}_t$ the corrupted sequence obtained by independently replacing each token of $\bm{o}$ with a mask token\footnote{The mask token $M$ serves as the absorbing state in the diffusion process.} $M \notin \mathcal{V}$ with probability depending on $t$.
The MDLM $p_\theta(o^{i} \mid \bm{o}_t, \bm{c})$ predicts the original token at position $i$ given the corrupted sequence $\bm{o}_t$ and the prompt $\bm{c}$, where $\theta$ denotes the model parameters.

\paragraph{Objective in pre-training.}
In pre-training, MDLMs maximize the Evidence Lower Bound\footnote{The summation in~(\ref{eq:mdlm_elbo}) runs only over positions that are masked in the corrupted sequence $\bm{o}_t$.} (ELBO)~\cite{NEURIPS2024_eb0b13cc}:
\begin{align}
  &\mathcal{L}_{\text{ELBO}}(\bm{o}; \theta)
  \notag\\
  &= \mathbb{E}_{t, \bm{o}_t}
  \!\left[
  \frac{1}{t}
  \sum_{i:\,o_{t}^{i}=M}
  \log p_\theta(o^{i} \mid \bm{o}_t, \bm{c})
  \right],
  \label{eq:mdlm_elbo}
\end{align}
where $t \sim \mathrm{Unif}((0, 1])$ and $\bm{o}_t \sim q(\cdot \mid \bm{o}, t)$; here $\mathrm{Unif}(X)$ denotes the uniform distribution over a set $X$.

\begin{figure}[t]
    \centering
    \includegraphics[width=\linewidth]{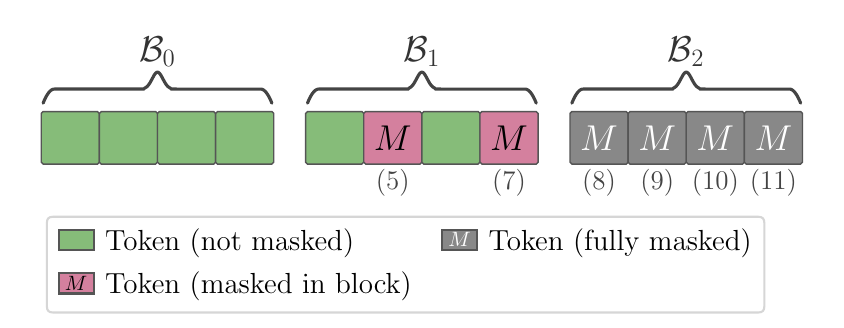}
    \caption{
Mask state during the generation of block $b=1$ for a response $\bm{o}$ of length $L=12$ ($C=3$, $B=4$). Block $\mathcal{B}_0$ is considered already generated, so its original tokens are retained. In this example, positions $5$ and $7$ within block $\mathcal{B}_1$ are masked ($\mathcal{M}_{t_1}=\{5,7\}$), and all positions in block $\mathcal{B}_2$ are masked. The full set of masked positions is $\mathcal{S}_1(\mathcal{M}_{t_1})=\{5,7,8,9,10,11\}$.
    }
    \label{fig:mask_state}
\end{figure}

\subsection{Blockwise generation}
\label{subsec:block_generation}
To achieve better generation quality, recent MDLMs adopt blockwise generation, which partitions a sequence into multiple blocks and generates them sequentially from left to right~\citep{ICLR2025_7ede97c3}.
When blockwise generation is used at inference time, it is desirable to compute the training objective based on the blockwise generation process as well.
This work considers reinforcement learning (RL) via policy optimization\footnote{Objectives for pre-training~\citep{ICLR2025_7ede97c3}~\citep{ICLR2026_2a2de7a4} and Supervised Fine-Tuning (SFT)~\citep{sun2025blockwise} have also been proposed.} and formulates a blockwise surrogate objective based on their implementation\footnote{\citet{ICLR2026_2a2de7a4} adopt a blockwise masking strategy in their implementation but do not explicitly formulate the corresponding objective (\url{https://github.com/facebookresearch/SPG}).}.

\paragraph{Notation.}
Let $B$ denote the block length and $C = L / B$ the number of blocks\footnote{For simplicity, we assume that $L$ is divisible by $B$.} when a response $\bm{o}$ of length $L$ is partitioned into blocks of length $B$.
Let $b\in\{0,\dots,C-1\}$ be the block index and $\mathcal{B}_b$ the set of token positions belonging to block $b$ ($|\mathcal{B}_b| = B$). We write $t_b\in(0,1]$ for the masking ratio within block $b$.
Let $\mathcal{M}_{t_b}\subseteq \mathcal{B}_b$ denote the set of positions masked with probability $t_b$ within block $b$.
During the generation of block $b$, all positions in ungenerated blocks $b' > b$ are also masked in addition to $\mathcal{M}_{t_b}$; we write $\mathcal{S}_b(\mathcal{M}_{t_b})$ for this extended mask set (see Appendix~\ref{app:mask_state} for the formal definition and Fig.~\ref{fig:mask_state} for an example).

\paragraph{Blockwise surrogate objective in RL.}
The number of masked tokens within block $b$ is $\max(1,\lfloor t_b B \rfloor)$, and we write $\Omega_b(t_b)$ for the set of all possible masked-position subsets\footnote{$\lfloor \cdot \rfloor$ denotes the floor function.} (see Appendix~\ref{app:omega_def} for the formal definition).
We also define the fraction of masked positions relative to the total sequence length as $\gamma_{b}(\mathcal{M}_{t_b}) = |\mathcal{S}_b(\mathcal{M}_{t_b})| / L$.
The blockwise surrogate objective is then defined as:
\begin{equation}
    \mathcal{L}_{\text{Block}}(\bm{o}\mid\bm{c};\theta):= \mathbb{E}_{b,t_b,\mathcal{M}_{t_b}}\left[\ell_{b,t_b,\mathcal{M}_{t_b}}(\bm{o},\bm{c};\theta)\right],
    \label{eq:block_obj}
\end{equation}
where
\begin{align}
    &\ell_{b,t_b,\mathcal{M}_{t_b}}(\bm{o},\bm{c};\theta)
    \notag\\
    &=
    \frac{1}{\gamma_{b}(\mathcal{M}_{t_b})}
    \sum_{i\in\mathcal{S}_b(\mathcal{M}_{t_b})}
    \log p_{\theta}\!\left(
    o^{i}\mid \bm{o}_{b,\mathcal{M}_{t_b}},\bm{c}
    \right),\label{eq:ell_b_t_b_M_t_b}
\end{align}
where $b\sim \mathrm{Unif}(\{0,\dots,C-1\})$, $t_b\sim \mathrm{Unif}((0, 1])$, and $\mathcal{M}_{t_b}\sim \mathrm{Unif}(\Omega_b(t_b))$.
Here, $\bm{o}_{b,\mathcal{M}_{t_b}}$ denotes the partially masked sequence obtained by replacing the positions in $\mathcal{S}_b(\mathcal{M}_{t_b})$ with $M$ (see Appendix~\ref{app:mask_state} for the formal per-position definition).
This gives the formal definition of (\ref{eq:block_obj_intro}).
The coefficient $1/\gamma_{b}(\mathcal{M}_{t_b})$ scales the token prediction loss by the number of masked positions, ensuring a consistent loss scale across different masking ratios.
We adopt Group Relative Policy Optimization (GRPO)~\cite{shao2024deepseekmath} for policy optimization and use $\mathcal{L}_{\text{Block}}$ as a surrogate for $\log \pi_\theta(\bm{o}^j\mid\bm{c})$ in the GRPO objective\footnote{
Here, $\bm{o}^j$ denotes the $j$-th of the $g$ responses $\{\bm{o}^j\}_{j=1}^g$ generated by the model for an input prompt $\bm{c}$.
} (see Appendix~\ref{app:grpo} for the formulation of GRPO).

\subsection{Instability of block selection}
\label{subsec:block_selection_instability}
Estimation of the blockwise surrogate objective relies on sampling blocks.
This work focuses on the finite-sample instability that arises when a small number of blocks are randomly selected.

\paragraph{Estimation of the blockwise surrogate objective.}
A finite-sample Monte Carlo estimate of $\mathcal{L}_{\text{Block}}$ in (\ref{eq:block_obj}) is obtained by $K$ independent samples:
\begin{equation}
    \hat{\mathcal{L}}_{\text{Rand}}(\bm{o}\mid\bm{c};\theta)
:= \frac{1}{K}\sum_{k=1}^{K} \ell_{b_k,t_k,\mathcal{M}_k}(\bm{o},\bm{c};\theta),
    \label{eq:rand_estimator}
\end{equation}
where $b_k \overset{\mathrm{i.i.d.}}{\sim} \mathrm{Unif}(\{0,\dots,C-1\})$, $t_k\overset{\mathrm{i.i.d.}}{\sim} \mathrm{Unif}((0, 1])$, and $\mathcal{M}_k\overset{\mathrm{i.i.d.}}{\sim} \mathrm{Unif}(\Omega_{b_k}(t_k))$.
However, in practice the number of blocks $C$ is typically small (e.g., $C = 2, 4$, or $8$).
Consequently, with a finite number of evaluations, the same block may be repeatedly sampled, while other blocks may never be selected.
As a result, the variability in block selection increases the instability of the estimate of $\mathcal{L}_{\text{Block}}$.

\begin{figure}[t]
    \centering
    \begin{subfigure}[t]{\linewidth}
        \centering
        \includegraphics[width=0.875\linewidth]{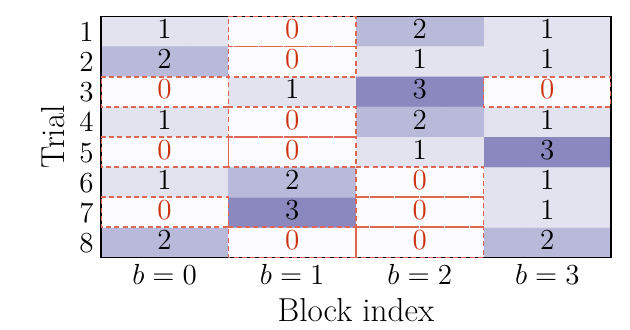}
        \caption{Per-block selection counts over 8 independent trials ($C{=}4, K{=}4$). Cells with 0 indicate missed blocks.}
        \label{fig:block_sampling_sim_a}
    \end{subfigure}\\[1.2ex]
    \begin{subfigure}[t]{\linewidth}
        \centering
        \includegraphics[width=0.875\linewidth]{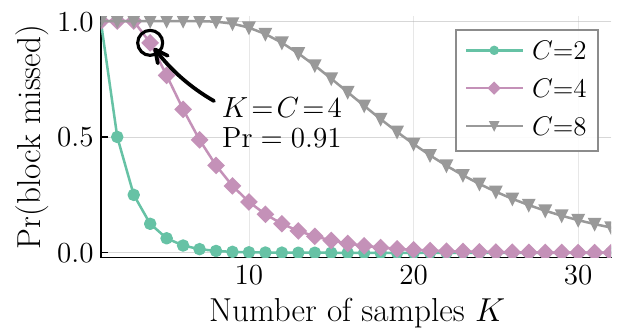}
        \caption{Probability that at least one block is never selected as a function of $K$ for varying $C$.}
        \label{fig:block_sampling_sim_b}
    \end{subfigure}
    \caption{Instability of random block sampling with a small number of blocks.
(a)~Some blocks receive zero selections after $K{=}4$ independent draws.
(b)~The miss probability increases rapidly with $C$
(see Appendix~\ref{app:miss_prob} for a formal derivation).}
    \label{fig:block_sampling_sim}
\end{figure}

\paragraph{Block sampling simulation.}
Figure~\ref{fig:block_sampling_sim} illustrates the finite-sample instability arising from random block selection.
Figure~\ref{fig:block_sampling_sim_a} shows the results of simulating 8 independent trials with $C = 4$ and $K = 4$; red-bordered cells indicate blocks that were never selected. In all trials, at least one block remains unselected.
Figure~\ref{fig:block_sampling_sim_b} shows the probability that at least one block is never selected for $C \in \{2, 4, 8\}$.
At $K = C = 4$, $\Pr = 0.91$, indicating that unselected blocks occur with high probability.

\section{Proposed Method}
We introduce a full-block estimator that evaluates all blocks rather than randomly selecting a subset, while keeping $\mathcal{L}_{\text{Block}}$ itself unchanged (see Fig.~\ref{fig:fig1}, top).

Since $\hat{\mathcal{L}}_{\text{Rand}}$ in (\ref{eq:rand_estimator}) randomly selects the block index $b$, the same block may be evaluated more than once.
The proposed full-block estimator evaluates every block exactly once and averages the results over all $C$ blocks:
\begin{align}
    &\hat{\mathcal{L}}_{\text{Full}}(\bm{o}\mid\bm{c};\theta)
:=
    \frac{1}{C}
    \sum_{b=0}^{C-1}
    \ell_{b,t_b,\mathcal{M}_{t_b}}(\bm{o},\bm{c};\theta),
    \label{eq:full_block_estimator}
\end{align}
where $t_b\sim\mathrm{Unif}((0, 1])$ and $\mathcal{M}_{t_b}\sim\mathrm{Unif}(\Omega_b(t_b))$ are sampled independently for each block $b$.

Indeed, $\hat{\mathcal{L}}_{\text{Full}}$ is an unbiased estimator of $\mathcal{L}_{\text{Block}}$:
\begin{align}
    &\mathbb{E}
    \left[
    \hat{\mathcal{L}}_{\text{Full}}(\bm{o}\mid\bm{c};\theta)
    \right] \notag \\
    &=
    \frac{1}{C}
    \sum_{b=0}^{C-1}
    \mathbb{E}_{t_b,\mathcal{M}_{t_b}}
    \left[
    \ell_{b,t_b,\mathcal{M}_{t_b}}(\bm{o},\bm{c};\theta)
    \right] \notag \\
    &=
    \mathcal{L}_{\text{Block}}(\bm{o}\mid\bm{c};\theta),
    \notag
\end{align}
where the last equality follows from the fact that each block $b$ is selected with probability $1/C$ under $b\sim \mathrm{Unif}(\{0,\dots,C-1\})$ in (\ref{eq:block_obj}).

\section{Theoretical Comparison of Objective Variances}
\label{sec:variance_reduction}
This section theoretically compares the variances of the proposed estimator $\hat{\mathcal{L}}_{\text{Full}}$ in (\ref{eq:full_block_estimator}) and the baseline $\hat{\mathcal{L}}_{\text{Rand}}$ in (\ref{eq:rand_estimator}) in terms of the within-group variance and the between-group variance.

\paragraph{Notation.}
For each block $b$, we define the expected loss and the variance of the loss within the block as
\begin{align}
    \mu_b
    &:=
    \mathbb{E}_{t_b,\mathcal{M}_{t_b}}
    \!\left[
    \ell_{b,t_b,\mathcal{M}_{t_b}}(\bm{o},\bm{c};\theta)
    \right], \notag \\
    \sigma_b^2
    &:=
    \mathrm{Var}_{t_b,\mathcal{M}_{t_b}}
    \!\left[
    \ell_{b,t_b,\mathcal{M}_{t_b}}(\bm{o},\bm{c};\theta)
    \right], \notag
\end{align}
respectively.
We further define the between-group variance, which measures the variability of the expected losses across blocks, and the within-group variance, which is the average of the variances within each block~\citep{https://doi.org/10.1111/j.1745-3984.1991.tb00363.x}:
\begin{align}
    \sigma_{\mathrm{between}}^2
    &:=
    \mathrm{Var}_{b^{\prime}}\!\left[\mu_{b^{\prime}}\right]
    =
    \frac{1}{C}\sum_{b=0}^{C-1}
    (\mu_b-\bar{\mu})^2, \label{eq:sig_between}\\
    \bar{\sigma}_{\mathrm{within}}^2
    &:=
    \mathbb{E}_{b^{\prime}}\!\left[\sigma_{b^{\prime}}^2\right]
    =
    \frac{1}{C}\sum_{b=0}^{C-1}
    \sigma_b^2, \label{eq:sig_within}
\end{align}
where ${b^{\prime}} \sim \mathrm{Unif}(\{0,\dots,C-1\})$ and $\bar{\mu}:= \frac{1}{C}\sum_{b=0}^{C-1}\mu_b$.

\paragraph{Variance of \texorpdfstring{$\hat{\mathcal{L}}_{\text{Rand}}$}{}.}
We derive the variance of each sample $\ell_{b_k,t_k,\mathcal{M}_k}$ in $\hat{\mathcal{L}}_{\text{Rand}}$ in (\ref{eq:rand_estimator}).
The randomness of each sample consists of two stages: the selection of the block index $b_k$ and the sampling of $(t_k,\mathcal{M}_k)$ within the selected block.
Applying the law of total variance~\citep{CaseBerg:01}, $\mathrm{Var}[X] = \mathrm{Var}[\mathbb{E}[X\mid Y]] + \mathbb{E}[\mathrm{Var}[X\mid Y]]$, with $X = \ell_{b_k,t_k,\mathcal{M}_k}$ and $Y = b_k$, we obtain
\begin{align}
    \mathrm{Var}\!\left[
    \ell_{b_k,t_k,\mathcal{M}_k}
    \right]
    \notag
    &=
    \mathrm{Var}_{b_k}\!\left[
    \mathbb{E}_{t_k,\mathcal{M}_k}\!\left[
    \ell_{b_k,t_k,\mathcal{M}_k} \mid b_k
    \right]
    \right] \notag \\
    &+
    \mathbb{E}_{b_k}\!\left[
    \mathrm{Var}_{t_k,\mathcal{M}_k}\!\left[
    \ell_{b_k,t_k,\mathcal{M}_k} \mid b_k
    \right]
    \right]
    \notag \\
    &= \mathrm{Var}_{b_k}\!\left[\mu_{b_k}\right]
    +\;\mathbb{E}_{b_k}\!\left[\sigma_{b_k}^2\right] \notag \\
    &=\sigma_{\mathrm{between}}^2+\bar{\sigma}_{\mathrm{within}}^2,
    \label{eq:var_ell}
\end{align}
where the last equality holds since $b_k \sim \mathrm{Unif}(\{0,\dots,C-1\})$.
Since the $K$ samples are mutually independent, from (\ref{eq:rand_estimator}) and (\ref{eq:var_ell}) we have
\begin{align}
    \mathrm{Var}\!\left[
    \hat{\mathcal{L}}_{\text{Rand}}
    \right]
    =
    \frac{1}{K}
    \!\left(
    \sigma_{\mathrm{between}}^2
    +
    \bar{\sigma}_{\mathrm{within}}^2
    \right).
    \label{eq:var_rand}
\end{align}

\paragraph{Variance of \texorpdfstring{$\hat{\mathcal{L}}_{\text{Full}}$}{}.}
In $\hat{\mathcal{L}}_{\text{Full}}$ in (\ref{eq:full_block_estimator}), the block index deterministically takes every value $b=0,\dots,C-1$ exactly once. The only random variables are $(t_b,\mathcal{M}_{t_b})$ within each block, and these are independent across blocks. From (\ref{eq:full_block_estimator}) and (\ref{eq:sig_within}),
\begin{align}
    \mathrm{Var}\!\left[
    \hat{\mathcal{L}}_{\text{Full}}
    \right]
    &=
    \frac{1}{C^2}
    \sum_{b=0}^{C-1}
    \sigma_b^2
    =
    \frac{\bar{\sigma}_{\mathrm{within}}^2}{C}.
    \label{eq:var_full}
\end{align}

\paragraph{Variance ratio.}
From (\ref{eq:var_rand}) and (\ref{eq:var_full}),
\begin{align}
    \frac{
    \mathrm{Var}
    \!\left[
    \hat{\mathcal{L}}_{\text{Full}}
    \right]
    }{
    \mathrm{Var}
    \!\left[
    \hat{\mathcal{L}}_{\text{Rand}}
    \right]
    }
    =
    \frac{K}{C}
    \cdot
    \frac{
    \bar{\sigma}_{\mathrm{within}}^2
    }{
    \sigma_{\mathrm{between}}^2
    +
    \bar{\sigma}_{\mathrm{within}}^2
    }.
    \label{eq:var_ratio}
\end{align}
Since the second factor in (\ref{eq:var_ratio}) is at most 1, the variance ratio is at most 1 whenever $K = C$ (i.e., the sample size matches the number of blocks).
That is, $\hat{\mathcal{L}}_{\text{Full}}$ remains unbiased for $\mathcal{L}_{\text{Block}}$ while eliminating the between-group variance $\sigma_{\mathrm{between}}^2$ arising from block selection, thereby achieving lower variance when $K = C$.

The variance ratio derived above in (\ref{eq:var_ratio}) is restricted to $K = C$ for simplicity. Appendix~\ref{app:variance_kneqc} extends the same analysis to arbitrary $K \geq 1$, and the resulting expression (\ref{eq:var_ratio_unified}) shows that the variance ratio is at most $1$. The variance reduction of our method ($\mathrm{Var}[\hat{\mathcal{L}}_{\text{Full}}] \leq \mathrm{Var}[\hat{\mathcal{L}}_{\text{Rand}}]$) therefore holds beyond the $K = C$ setting.

\begin{table*}[t]
    \small
    \centering
    \setlength{\tabcolsep}{3pt}
    \begin{tabular}{l c cccccc}
        \toprule
        & & \multicolumn{6}{c}{\textbf{Benchmark}} \\
        \cmidrule(lr){3-8}
        \textbf{Method} & \textbf{Blockwise} & GSM8K & MATH-500 & Countdown & Sudoku & HumanEval & Avg. \\
        \midrule
        LLaDA & \texttimes & 76.27 & 31.80 & 19.14 & 26.56 & 27.44 & 36.24 \\
        LLaDA-1.5 & \texttimes & 78.70 & 33.00 & 18.36 & 23.64 & 32.93 & 37.33 \\
        D1 & \texttimes & \secondbest{81.43} & 36.00 & 36.72 & 13.21 & 32.93 & 40.06 \\
        WD1 & \texttimes & 81.20 & 34.60 & 44.14 & 62.25 & 28.05 & 50.05 \\
        \midrule
        SPG & \checkmark & 80.29 & \secondbest{36.20} & \textbf{70.31}\gdiff{3.90} & \secondbest{72.73} & 29.88 & \secondbest{57.88} \\
        \rowcolor{okabeorange!8}
        $\hat{\mathcal{L}}_{\mathrm{Rand}}$ & \checkmark & 76.20 & 34.40 & 44.14 & 61.20 & \secondbest{34.76} & 50.14 \\
        \midrule
        \rowcolor{okabeblue!8}
        $\hat{\mathcal{L}}_{\mathrm{Full}}$ (ours) & \checkmark & \textbf{82.26} & \textbf{38.00} & \secondbest{66.41} & \textbf{91.39} & \textbf{36.59} & \textbf{62.93} \\[-1pt]
        \rowcolor{okabeblue!8}
        & & \rdiff{6.06}\,/\,\gdiff{0.83} & \rdiff{3.60}\,/\,\gdiff{1.80} & \rdiff{22.27} & \rdiff{30.19}\,/\,\gdiff{18.66} & \rdiff{1.83}\,/\,\gdiff{1.83} & \rdiff{12.79}\,/\,\gdiff{5.05} \\
        \bottomrule
    \end{tabular}
    \caption{
    Performance comparison across benchmarks. Higher scores indicate better performance. LLaDA denotes the base model (LLaDA-8B-Instruct) before fine-tuning. The \textbf{Blockwise} column indicates methods that use the blockwise surrogate objective (\ref{eq:block_obj}). \textbf{Bold} denotes the highest accuracy in each column, and $\dagger$ denotes the second highest. {\color{okabered!80!black}$\blacktriangle$} indicates the improvement of $\hat{\mathcal{L}}_{\mathrm{Full}}$ over $\hat{\mathcal{L}}_{\mathrm{Rand}}$, and {\color{okabegreen!80!black}$\uparrow$} indicates the improvement of the highest accuracy over the $\dagger$ value.
    }
    \label{tab:main_results_styled}
\end{table*}

\section{Experimental Setup}\label{sec:main-exp-setup}
This section evaluates the proposed full-block estimator $\hat{\mathcal{L}}_{\text{Full}}$ in (\ref{eq:full_block_estimator}) on five benchmarks across three categories: mathematics, planning, and coding.
Specifically, we assess (i) whether the difference in estimator alone leads to a performance gap when the sample size matches the number of blocks ($\hat{\mathcal{L}}_{\text{Full}}$ in (\ref{eq:full_block_estimator}) vs.\ $\hat{\mathcal{L}}_{\text{Rand}}$ in (\ref{eq:rand_estimator})), and (ii) whether our method achieves competitive performance compared with existing MDLM post-training methods.

\subsection{Baselines}
We use LLaDA-8B-Instruct~\citep{NEURIPS2025_48b383b2} as the base model.
The compared methods are LLaDA-1.5~\citep{zhu-etal-2026-llada}, D1~\citep{NEURIPS2025_52190a03}, WD1~\citep{ICLR2026_f381114c}, SPG~\citep{ICLR2026_2a2de7a4}, and $\hat{\mathcal{L}}_{\text{Rand}}$ in (\ref{eq:rand_estimator}).\footnote{LLaDA-8B-Instruct is one of the largest publicly available instruction-tuned MDLMs, and the compared methods are also evaluated on LLaDA-family MDLMs; we therefore adopt it as the base model in this work. LLaDA-1.5, D1, WD1, and SPG are all standard baselines used in prior work on MDLM post-training.}
In particular, the comparison between $\hat{\mathcal{L}}_{\text{Rand}}$ in (\ref{eq:rand_estimator}), which samples the block index independently and uniformly, and our method $\hat{\mathcal{L}}_{\text{Full}}$ in (\ref{eq:full_block_estimator}) serves as an ablation that isolates the effect of the estimator while keeping the objective fixed.
See Section~\ref{sec:related_work} for details on each method.

\subsection{Tasks}\label{sec:tasks}
We use five benchmarks spanning the following three task categories.
(i)~\textbf{Mathematics}: GSM8K~\citep{cobbe2021training} and MATH-500~\citep{ICLR2024_aca97732}.
(ii)~\textbf{Planning}: Countdown~\citep{tinyzero} and Sudoku~\citep{sudoku}.
(iii)~\textbf{Coding}: HumanEval~\citep{chen2021codex}. The training data for the coding task are taken from \citet{open_r1_verifiable_coding_problems}.

\subsection{Training Details}

\paragraph{Model settings.}
Our method and all compared methods are trained with the same hyperparameter settings.
In all experiments, we fine-tune the base model (LLaDA-8B-Instruct) using QLoRA~\citep{NEURIPS2023_1feb8787} (4-bit NF4 quantization, LoRA rank $r=128$; $r=64$ for the coding task).
We use the AdamW optimizer~\cite{loshchilov2018decoupled} with a learning rate of $1.5 \times 10^{-6}$ (constant schedule with warmup), and all tasks are trained for 4,000 steps.

\paragraph{Generation settings.}
During training, the generation length is $L=256$ ($512$ for the coding task) and the block length is $B = 64$ ($128$ for the coding task), yielding $C = 4$ blocks for all tasks.
The number of diffusion steps is set to half the generation length, corresponding to an average of 2 tokens unmasked per step.

\paragraph{Objective settings.}
We use GRPO~\cite{shao2024deepseekmath} for policy optimization with task-specific rule-based verifiers as reward functions (see Appendix~\ref{sec:reward_functions} for more details).
$\hat{\mathcal{L}}_{\text{Full}}$, $\hat{\mathcal{L}}_{\text{Rand}}$, and SPG all employ blockwise surrogate objectives, and we set the number of Monte Carlo samples to $K = C = 4$ for all three methods.
Under this setting, $\hat{\mathcal{L}}_{\text{Full}}$ evaluates each block exactly once, while SPG and $\hat{\mathcal{L}}_{\text{Rand}}$ independently sample $(b_k,t_k,\mathcal{M}_k)$ $K$ times.
D1 and WD1 do not use the blockwise surrogate objective, so this sample size is not applicable.

See Appendix~\ref{app:training_details} for further details.

\subsection{Evaluation}
Following prior work~\citep{ICLR2026_2a2de7a4,NEURIPS2025_52190a03}, the evaluation settings for each task are as follows.
(i)~\textbf{Mathematics \& Planning}: GSM8K and MATH-500 are evaluated in a 0-shot setting using exact match accuracy between the generated answer and the ground-truth label.
For the mathematics tasks, we extract the expression inside \texttt{\textbackslash boxed\{\}}, normalize the LaTeX representation, and check for mathematical equivalence.
For Countdown and Sudoku, correctness is determined by whether the generated answer matches the ground truth. Sudoku is evaluated in a 2-shot setting.
(ii)~\textbf{Coding}: HumanEval is evaluated in a 2-shot setting using Pass@1 (the fraction of problems for which a single generation passes all test cases).

We evaluate each task using the checkpoint from the final training step.
See Appendix~\ref{app:eval_details} for details on inference parameters.

\section{Results}\label{sec:results}
Table~\ref{tab:main_results_styled} presents the results.
Our method achieves the highest accuracy on four out of five benchmarks and outperforms SPG, the strongest baseline, by 5.05 points in the simple average across all five benchmarks.

We first compare $\hat{\mathcal{L}}_{\text{Full}}$ in (\ref{eq:full_block_estimator}) and $\hat{\mathcal{L}}_{\text{Rand}}$ in (\ref{eq:rand_estimator}) as an ablation that isolates the effect of the estimator.
$\hat{\mathcal{L}}_{\text{Full}}$ outperforms $\hat{\mathcal{L}}_{\text{Rand}}$ on all five benchmarks, with an average improvement of 12.79 points.
This result shows that, even with the same surrogate objective, the design of the estimator affects the final task performance in the finite-sample regime.

Our method also demonstrates consistent performance compared with all existing methods.
On the mathematics tasks, our method achieves the highest accuracy on both GSM8K (82.26) and MATH-500 (38.00).
On Sudoku, the performance gap among methods is large; our method (91.39) outperforms the strongest baseline, SPG (72.73), by 18.66 points.
On HumanEval, a coding task, our method (36.59) also outperforms all baselines.
On the other hand, on Countdown, our method (66.41) outperforms $\hat{\mathcal{L}}_{\text{Rand}}$ (44.14) by 22.27 points but does not surpass SPG (70.31).
SPG combines (a) the sandwiched objective and (b) random block sampling. Our method does not use (a), and instead improves (b).
The gap on Countdown may therefore be attributable to the effect of (a).

We examine whether these results depend on the specific setting $K = C = 4$ through the additional experiments in Appendix~\ref{app:k_lt_c_empirical} ($K \in \{2, 6, 8\}$ with $C = 4$ fixed) and Appendix~\ref{app:c_scalability} (doubling to $C = 8$). Our method matches or outperforms $\hat{\mathcal{L}}_{\text{Rand}}$ in every setting.\footnote{The improvements over $\hat{\mathcal{L}}_{\text{Rand}}$ are as follows: at $K = 2$, +6.66 points on the five-dataset average; at $K = 8$, +0.33 points on the GSM8K/MATH-500 average; and at $C = 8$, +2.92 points on the GSM8K/MATH-500 average.}
Note that within the settings we examined, the gain of our method is largest in the low-sample regime $K = C$ and generally shrinks as the sample size $K$ grows beyond the number of blocks $C$ (Appendix~\ref{app:k_lt_c_empirical}).

The next section analyzes the difference between our method and the existing estimator more directly from the perspectives of estimation quality and variance.

\section{Empirical Comparison of Objective Estimates}
\label{sec:analysis}
This section quantitatively compares the statistical properties of $\hat{\mathcal{L}}_{\text{Full}}$ in (\ref{eq:full_block_estimator}) and $\hat{\mathcal{L}}_{\text{Rand}}$ in (\ref{eq:rand_estimator}).
To this end, we first describe the experimental setup, then evaluate the overall trend of estimation quality using the Spearman rank correlation and RMSE with respect to a high-precision reference estimate.
Finally, we verify the theoretical variance ratio derived in (\ref{eq:var_ratio}) using empirical measurements.

\subsection{Setup}
For the five benchmarks described in Section~\ref{sec:tasks}, we extract $n=1{,}000$ prompts from the training data and generate one response per prompt using LLaDA-8B-Instruct\footnote{We use the base model rather than the trained model from Section~\ref{sec:results} so that the statistical properties of $\hat{\mathcal{L}}_{\text{Full}}$ and $\hat{\mathcal{L}}_{\text{Rand}}$ are compared under the same conditions.}, with the number of blocks set to $C = 4$.
Since $\hat{\mathcal{L}}_{\text{Full}}$ evaluates each block exactly once, we set $K = C = 4$ for $\hat{\mathcal{L}}_{\text{Rand}}$ to match the sample size.

In what follows, we refer to $\hat{\mathcal{L}}_{\text{Rand}}$ computed with sample size $K = 1024$ as the high-precision reference estimate $\mathcal{L}_{\mathrm{ref}}$.
Although this is not the exact true value, we regard it as an approximate reference with sufficient precision for the comparisons in this section\footnote{For example, on GSM8K the Spearman rank correlation between the estimates at $K = 512$ and $K = 1024$ is $0.994$, indicating that the ranking is nearly stable and further increasing $K$ has little effect.}.

\subsection{Estimation Quality}
\label{subsec:estimation_quality}

\begin{figure}[t]
    \small
    \centering
        \begin{subfigure}[t]{0.875\linewidth}
        \centering
        \includegraphics[width=\linewidth]{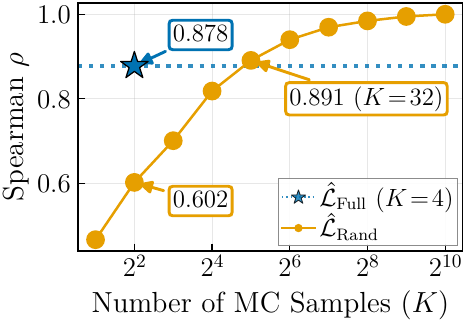}
        \caption{Spearman $\rho$}
        \label{fig:mc_accuracy_trend_rho}
    \end{subfigure}\\[1.2ex]
    \begin{subfigure}[t]{0.875\linewidth}
        \centering
        \includegraphics[width=\linewidth]{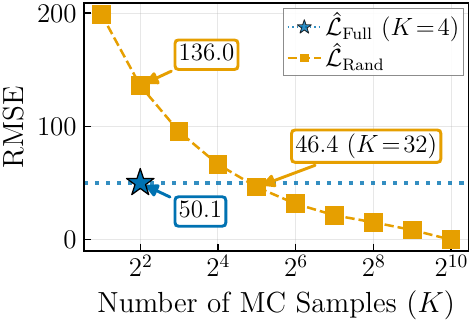}
        \caption{RMSE}
        \label{fig:mc_accuracy_trend_rmse}
    \end{subfigure}
    \caption{
Comparison of estimation quality for $\mathcal{L}_{\text{Block}}$ in (\ref{eq:block_obj}) between $\hat{\mathcal{L}}_{\text{Full}}$ in (\ref{eq:full_block_estimator}) and $\hat{\mathcal{L}}_{\text{Rand}}$ in (\ref{eq:rand_estimator}) on GSM8K ($C = 4$).
$\hat{\mathcal{L}}_{\text{Rand}}$ is computed with varying sample size $K$.
The estimate at $K = 1024$ is used as the high-precision reference estimate $\mathcal{L}_{\mathrm{ref}}$ for computing Spearman $\rho$ and RMSE.
Higher $\rho$ and lower RMSE indicate better estimation quality.
The star indicates the result of $\hat{\mathcal{L}}_{\text{Full}}$ at $K = C = 4$, which achieves estimation quality roughly comparable to $\hat{\mathcal{L}}_{\text{Rand}}$ at $K \approx 32$.
    }
    \label{fig:mc_accuracy_trend}
\end{figure}

We evaluate how close the estimates of $\mathcal{L}_{\text{Block}}$ in (\ref{eq:block_obj}) are to the reference value $\mathcal{L}_{\mathrm{ref}}$ under a finite sample size.
As evaluation metrics, we use the Spearman rank correlation coefficient $\rho$ and the Root Mean Squared Error\footnote{The RMSE for $n$ samples is defined as:
$$
\mathrm{RMSE}
=
\sqrt{
\frac{1}{n}
\sum_{i=1}^{n}
\left(
\hat{\mathcal{L}}_i
-
\mathcal{L}_{\mathrm{ref},i}
\right)^2
},
$$
where $\hat{\mathcal{L}}_i$ and $\mathcal{L}_{\mathrm{ref},i}$ are the estimate ($\hat{\mathcal{L}}_{\text{Full}}$ or $\hat{\mathcal{L}}_{\text{Rand}}$) and the reference value for the $i$-th sample, respectively.} (RMSE) with respect to the reference value.
The rank correlation measures how well the relative ordering of the estimates is preserved, and the RMSE measures the magnitude of the estimation error.
For $\hat{\mathcal{L}}_{\text{Rand}}$, we compute estimates with sample size $K = 2, 4, 8, \ldots, 1024$ and compare them against $\mathcal{L}_{\mathrm{ref}}$.

Figure~\ref{fig:mc_accuracy_trend} shows the results on GSM8K.
At sample size $K = 4$, $\hat{\mathcal{L}}_{\text{Full}}$ achieves $\rho = 0.878$ and $\mathrm{RMSE} = 50.1$, substantially outperforming $\hat{\mathcal{L}}_{\text{Rand}}$ ($\rho = 0.602$, $\mathrm{RMSE} = 136.0$).
This level of accuracy is comparable to $\hat{\mathcal{L}}_{\text{Rand}}$ at $K \approx 32$, demonstrating that merely eliminating the random sampling of the block index substantially improves the estimation quality in the finite-sample regime.

Similar trends are observed on the other datasets (Appendices~\ref{subsec:error_stats}, \ref{subsec:per_dataset_details}); $\hat{\mathcal{L}}_{\text{Full}}$ further exhibits substantially lower tail probability of estimation errors than $\hat{\mathcal{L}}_{\text{Rand}}$ (Appendix~\ref{subsec:tail_probability}).

\begin{table}[t]
    \small
    \centering
    \begin{tabular}{ccccc}
        \toprule
        GSM8K & MATH & Countdown & Sudoku & Code \\
        \midrule
        0.122 & 0.119 & 0.095 & 0.076 & 0.107 \\
        \bottomrule
    \end{tabular}
    \caption{
    Variance ratio $\mathrm{Var}[\hat{\mathcal{L}}_{\text{Full}}] \mathbin{/} \mathrm{Var}[\hat{\mathcal{L}}_{\text{Rand}}]$ in (\ref{eq:var_ratio}), evaluated numerically for each sample and averaged within each dataset ($K = C = 4$). Smaller values indicate a greater variance reduction effect of $\hat{\mathcal{L}}_{\text{Full}}$.
    }
    \label{tab:variance_ratio}
\end{table}

\subsection{Quantitative Evaluation of the Variance Ratio}
\label{subsec:variance_ratio}

We quantitatively evaluate the variance ratio derived in (\ref{eq:var_ratio}).
We then also discuss the variance ratio at the policy-gradient level.

\paragraph{Estimator level.}
For each of the $C = 4$ blocks, we independently sample the diffusion timestep 50 times\footnote{For each block $b$, we independently sample $t_b$ and $\mathcal{M}_{t_b}$ 50 times.} and compute $\ell_{b,t_b,\mathcal{M}_{t_b}}(\bm{o},\bm{c};\theta)$ in (\ref{eq:ell_b_t_b_M_t_b}).
From these sampled values, we compute $\sigma_{\mathrm{between}}^2$ in (\ref{eq:sig_between}) and $\bar{\sigma}_{\mathrm{within}}^2$ in (\ref{eq:sig_within}).
We then evaluate the right-hand side of (\ref{eq:var_ratio}) with $K = C = 4$ for each sample.
Table~\ref{tab:variance_ratio} reports the average of these values within each dataset.
The variance ratio ranges from $0.08$ to $0.12$ across all datasets, indicating that the variance of $\hat{\mathcal{L}}_{\text{Full}}$ in (\ref{eq:var_full}) is approximately $1/10$ of the variance of $\hat{\mathcal{L}}_{\text{Rand}}$ in (\ref{eq:var_rand}).

As shown in (\ref{eq:var_ratio}), when $K = C$, the difference between the variance of $\hat{\mathcal{L}}_{\text{Full}}$ and that of $\hat{\mathcal{L}}_{\text{Rand}}$ lies in the presence or absence of the between-group variance $\sigma_{\mathrm{between}}^2$ arising from block selection.
The results in Table~\ref{tab:variance_ratio} show that $\sigma_{\mathrm{between}}^2$ accounts for the majority of $\mathrm{Var}[\hat{\mathcal{L}}_{\text{Rand}}]$.
By eliminating this component, $\hat{\mathcal{L}}_{\text{Full}}$ provides a more stable surrogate estimate at the estimator level.

\paragraph{Policy-gradient level.}
We also examine how far this variance reduction carries over to the policy gradient, which is what actually updates the policy. On GSM8K and MATH-500 at mid-training checkpoints, the estimator-level variance ratio is 0.10--0.16 (Appendix~\ref{app:variance_ratio_post}), whereas the policy-gradient variance ratio is 0.79--0.96 (Appendix~\ref{app:gradient_variance_argument}), so the accuracy gains in Section~\ref{sec:results} are not necessarily due to reduced gradient variance. Unlike $\hat{\mathcal{L}}_{\text{Rand}}$, which is likely to leave some blocks unselected at each step (Section~\ref{subsec:block_selection_instability}), $\hat{\mathcal{L}}_{\text{Full}}$ delivers a learning signal to every block at every step, which we consider another source of its advantage. We do not disentangle how much each of the two factors, estimator-level variance reduction and the learning signal delivered to every block, contributes to the accuracy difference.

\section{Related Work}
\label{sec:related_work}

\paragraph{Masked Diffusion Language Models.}
Discrete diffusion models originated from D3PM~\citep{NEURIPS2021_958c5305} and have since evolved through SEDD~\citep{pmlr-v235-lou24a}, MDLM~\citep{NEURIPS2024_eb0b13cc}, and MD4~\citep{NEURIPS2024_bad233b9}.
LLaDA~\citep{NEURIPS2025_48b383b2} scaled masked diffusion models to 8B parameters, demonstrating performance comparable to auto-regressive language models and introducing efficient inference via blockwise generation.
This work targets the post-training based on blockwise generation and the surrogate objective adopted by LLaDA.

\paragraph{Blockwise Training for MDLM.}
For pre-training, \citet{ICLR2025_7ede97c3} proposed block diffusion, which performs discrete diffusion within blocks and generates across blocks auto-regressively, and introduced a noise schedule that minimizes gradient variance.
For SFT, \citet{sun2025blockwise} proposed computing the loss by selecting a single active block, and \citet{ICLR2026_d0865cbe} proposed converting ARLMs to block diffusion models via a complementary masking strategy.

\paragraph{Reinforcement Learning for MDLM.}
Within the same GRPO + blockwise surrogate framework as our method, diffu-GRPO~\citep{NEURIPS2025_52190a03} proposes one-step log-likelihood estimation via a mean-field approximation, WD1~\citep{ICLR2026_f381114c} proposes a ratio-free objective, and SPG~\citep{ICLR2026_2a2de7a4} proposes a sandwiched policy gradient.
VRPO~\citep{zhu-etal-2026-llada} performs an MDLM variance analysis in the preference-based setting\footnote{VRPO performs antithetic sampling for a preference-score estimator (the ELBO difference on the DPO preference pair $(y_w, y_l)$), which presupposes the existence of $(y_w, y_l)$ and a reference policy $\pi_{\text{ref}}$; an estimator-level comparison between VRPO and our method is therefore not feasible. Likewise, objective-level prior work such as SPG cannot be applied to VRPO.}.
RL methods compatible with blockwise generation have also been proposed, including DiRL~\citep{zhu2025dirl}, GDPO~\citep{ICLR2026_0428b72c}, and StableDRL~\citep{zhong2026stabilizing}.

\paragraph{Relation to Classical Sampling Techniques.}
When $K = C$, $\hat{\mathcal{L}}_{\text{Full}}$ is equivalent to stratified sampling with one sample per stratum~\citep{mcbook}. The contribution of this work is to introduce the perspective of Monte Carlo estimator design into the blockwise generation structure of MDLM. See Appendix~\ref{app:related_extension} for discussion, including the extension to importance sampling.
We also derive the variance ratio for general $K$, including $K \neq C$ (Appendix~\ref{app:variance_kneqc}). Moreover, our method can be viewed as a form of Rao--Blackwellization, in that it computes the expectation over the block index exactly. The variance that our method targets arises from block selection and differs from the variance targeted by the antithetic sampling of VRPO~\citep{zhu-etal-2026-llada}.

While these prior studies focus on designing the surrogate objective, approximating the likelihood, or improving the policy optimization rule, this work explicitly examines the missing coverage and the variance arising from block index selection in the Monte Carlo estimator while keeping the blockwise surrogate objective fixed.

\section{Conclusion}
This work proposed a full-block estimator $\hat{\mathcal{L}}_{\text{Full}}$ that evaluates every block exactly once to address the missing coverage and the estimation variance caused by random block-index sampling in the Monte Carlo estimation of the blockwise surrogate objective for MDLM post-training.
The proposed estimator remains unbiased while eliminating the between-group variance from block selection, and it achieved the highest accuracy on four out of five benchmarks, outperforming the random block sampling baseline by an average of 12.79 points.
These results indicate that the design of the estimator is also important for improvement in MDLM post-training, in addition to the design of the surrogate objective and the policy optimization rule.

\section*{Limitations}
\begin{itemize}
    \item Due to computational resource constraints, we did not conduct experiments with increased training steps or GRPO group size $g$.
    \item Because our estimator changes only how the block index is selected, it is in principle applicable to other policy optimization methods, such as PPO~\cite{schulman2017proximal} and DPO~\cite{NEURIPS2023_a85b405e}, as long as they use a blockwise surrogate objective. However, we have not verified its effect in combination with these methods.
    \item Although LLaDA-8B-Instruct is one of the largest publicly available instruction-tuned MDLMs (Section~\ref{sec:main-exp-setup} footnote), generalization to other model sizes and architectures as well as to block counts beyond $C \in \{4, 8\}$ requires further investigation.
    \item Our evaluation is limited to three task categories (mathematics, planning, and coding) and the effectiveness on other task categories such as open-ended natural language generation and summarization has not been confirmed.
    \item We evaluate the accuracy of every method only at its final checkpoint and do not compare accuracy at intermediate checkpoints.
    \item Since $\hat{\mathcal{L}}_{\text{Full}}$ evaluates all $C$ blocks in parallel, GPU VRAM usage scales linearly with $C$. In our experiments, $C = 4$ was small enough to avoid this issue. For larger $C$, this $O(C)$ overhead is specific to the parallel-batch implementation and can be relaxed by sequentially accumulating gradients over blocks, which reduces peak VRAM to a single-block footprint at the cost of $C\times$ wall-clock time per iteration; a systematic memory-vs-time trade-off study is left for future work.
    \item We have not disentangled how much each of the two factors, estimator-level variance reduction and the learning signal delivered to every block, contributes to the accuracy difference between our method and $\hat{\mathcal{L}}_{\text{Rand}}$.
    \item Our full-block evaluation targets the block index, which is discrete and takes only a few values. Extending the same idea to continuous or high-dimensional random variables, such as diffusion timesteps and masked positions, requires approximation and is left for future work.
\end{itemize}

\section*{Acknowledgments}
This work was partially supported by JSPS KAKENHI JP22H05106, JP23K28045, JP26H02495 (HS), JP25K24366 (HY), JST CREST JPMJCR21N3 (HS), JST Moonshot JPMJMS263I (HS), and JST BOOST JPMJBS2407 (YH).

\bibliography{citation_checked}

\appendix

\section{Experimental Details}
\label{sec:experimental_details}
This section provides detailed experimental settings. All methods evaluated in this study---D1, WD1, SPG, $\hat{\mathcal{L}}_{\text{Rand}}$, and $\hat{\mathcal{L}}_{\text{Full}}$---share the following common configuration.

\subsection{Training Details}
\label{app:training_details}
We describe the detailed training configuration.
We use LLaDA-8B-Instruct~\cite{NEURIPS2025_48b383b2} as the base model, which is one of the largest publicly available instruction-tuned MDLMs.
For memory efficiency, we apply QLoRA (4-bit NF4) quantization~\cite{NEURIPS2023_1feb8787}.
All experiments are conducted using a PyTorch-based framework with Accelerate and DeepSpeed (ZeRO Stage 2)~\cite{9355301} on two NVIDIA GPUs (RTX A6000 or RTX 6000 Ada Generation), with bfloat16 precision.
We set the random seed to $42$ for all experiments.
We follow the prompt templates of \citet{ICLR2026_2a2de7a4} for all tasks.

\paragraph{Hyperparameters.}
Table~\ref{tab:common_hyperparams} lists the optimization, LoRA, and training hyperparameters, and Table~\ref{tab:diffusion_train} lists the diffusion generation settings during training.
All methods share these hyperparameters.\footnote{Because of limited computational resources, our training settings, such as the batch size, differ from those of the SPG paper~\citep{ICLR2026_2a2de7a4}. Our numbers are therefore not directly comparable to the values reported in that paper, and we retrain all methods under this single setting for comparison.}

\begin{table}[t]
    \centering
    \small
    \begin{tabular}{lc}
        \toprule
        \textbf{Hyperparameter} & \textbf{Value} \\
        \midrule
        \multicolumn{2}{c}{\textit{Optimization}} \\
        Optimizer & AdamW \\
        Learning rate & $1.5 \times 10^{-6}$ \\
        LR scheduler & constant\_with\_warmup \\
        Weight decay & $0.1$ \\
        Max gradient norm & $0.2$ \\
        \midrule
        \multicolumn{2}{c}{\textit{LoRA (QLoRA, 4-bit NF4)}} \\
        Rank ($r$) & 128\textsuperscript{\ddag} \\
        Alpha ($\alpha$) & 64\textsuperscript{\ddag} \\
        Dropout & 0.05 \\
        Target modules & All linear layers\textsuperscript{\dag} \\
        \midrule
        \multicolumn{2}{c}{\textit{Training}} \\
        Number of GPUs & 2 \\
        Per-device batch size & 2\textsuperscript{\ddag} \\
        Gradient accum.\ steps & 3 \\
        Num.\ generations ($g$) & 4\textsuperscript{\ddag} \\
        Num.\ iterations ($\mu$\textsuperscript{\dag\dag}) & 4 \\
        Epsilon (clipping) & 0.5 \\
        Beta (KL coeff.) & 0.0 \\
        Max steps & 4000 \\
        Max prompt length & 256\textsuperscript{\ddag\,\S} \\
        Max completion length ($L$) & 256\textsuperscript{\ddag} \\
        \bottomrule
        \multicolumn{2}{l}{
            \textsuperscript{\dag}\footnotesize
            \texttt{q,k,v,o,up,down,gate\_proj}.
        } \\
        \multicolumn{2}{l}{
            \textsuperscript{\dag\dag}\footnotesize
            GRPO iteration numbers.
        } \\
        \multicolumn{2}{l}{
            \textsuperscript{\ddag}\footnotesize
            Code task uses $r=64$, $\alpha=32$, $g=2$, $L=512$.
        } \\
        \multicolumn{2}{l}{
            \footnotesize
            Code: per-device batch size 1, max prompt length 1800.
        } \\
        \multicolumn{2}{l}{
            \textsuperscript{\S}\footnotesize
            Sudoku: max prompt length 1000 (2-shot prompts).
        }
    \end{tabular}
    \caption{Common training hyperparameters.}
    \label{tab:common_hyperparams}
\end{table}

\begin{table}[t]
    \centering
    \small
    \begin{tabular}{lcc}
        \toprule
        \textbf{Parameter} & \textbf{Default} & \textbf{Code} \\
        \midrule
        Temperature & 0.9\textsuperscript{*} & 0.9 \\
        Block length & 64 & 128 \\
        Diffusion steps & 128 & 256 \\
        Generation length & 256 & 512 \\
        Remasking strategy & \multicolumn{2}{c}{low\_confidence} \\
        \bottomrule
    \end{tabular}
    \caption{Diffusion generation settings during training. Default applies to GSM8K, MATH, Countdown and Sudoku. \textsuperscript{*}Sudoku uses 0.3.}
    \label{tab:diffusion_train}
\end{table}

\paragraph{Datasets.}
Table~\ref{tab:datasets} lists the training and evaluation data sources for each task.
We select five tasks that span diverse domains: mathematical reasoning (GSM8K, MATH), planning and search (Countdown, Sudoku), and code generation (Code).
GSM8K~\cite{cobbe2021training} consists of grade-school multi-step arithmetic reasoning problems.
For MATH, we use \texttt{ankner/math-500} as the training data, which contains competition-level problems in algebra, geometry, number theory, and other topics.
Countdown~\cite{tinyzero} requires combining given numbers with arithmetic operations to reach a target value.
Sudoku~\cite{ICLR2026_2a2de7a4} is a $4 \times 4$ puzzle that requires filling in empty cells while satisfying row, column, and box constraints.
For Code, we adopt HumanEval~\cite{chen2021codex} as the evaluation benchmark; however, since HumanEval does not provide a training split, we use Verifiable Coding Problems from Open-R1~\cite{openr1} as the training data, which consists of Python programming problems with automatic test-case verification.
Because the Code task requires a longer generation length ($512$ tokens), we adjust the LoRA rank and the number of generations to accommodate memory constraints (Table~\ref{tab:common_hyperparams}).

\begin{table*}[t]
    \centering
    \small
    \begin{tabular}{lll}
        \toprule
        \textbf{Task} & \textbf{Train Source}
            & \textbf{Eval Source} \\
        \midrule
        GSM8K
            & \texttt{gsm8k} (train)~\cite{cobbe2021training}
            & \texttt{gsm8k} (test) \\
        MATH
            & \texttt{ankner/math-500}\tabfnmark{math}
            & \texttt{MATH-500}~\cite{ICLR2024_aca97732} \\
        Countdown
            & \texttt{Jiayi-Pan/Countdown-Tasks-3to4}\tabfnmark{countdown}
            & \texttt{Jiayi-Pan/Countdown-Tasks-3to4} \\
        Sudoku
            & Custom 4$\times$4 (train)~\cite{ICLR2026_2a2de7a4}
            & Custom 4$\times$4 (test) \\
        Code
            & Verifiable Coding Problems\tabfnmark{code}
            & HumanEval~\cite{chen2021codex} \\
        \bottomrule
    \end{tabular}
    \caption{Datasets used for training and evaluation.}
    \label{tab:datasets}
\end{table*}
\afterpage{\afterpage{%
  \tabfntext{math}{\url{https://huggingface.co/datasets/ankner/math-500}}%
  \tabfntext{countdown}{\url{https://huggingface.co/datasets/Jiayi-Pan/Countdown-Tasks-3to4}}%
  \tabfntext{code}{\url{https://huggingface.co/datasets/open-r1/verifiable-coding-problems-python_decontaminated-tested}}}}

\paragraph{Reward Functions.}
The design of the reward function for each task is described in Appendix~\ref{sec:reward_functions}.

\subsection{Evaluation Details}
\label{app:eval_details}

We describe the inference parameters and benchmark details used for evaluation.

\paragraph{Generation Hyperparameters.}
At inference time, we use greedy decoding (Temperature $= 0.0$) for reproducibility. We adopt the \texttt{low\_confidence} remasking strategy~\cite{NEURIPS2025_48b383b2}, which preferentially remasks low-confidence tokens. For a generation length $L$, the number of diffusion steps is uniformly set to $L/2$. Table~\ref{tab:eval_common} lists the common settings.

\begin{table}[t]
    \centering
    \small
    \begin{tabular}{lc}
        \toprule
        \textbf{Parameter} & \textbf{Value} \\
        \midrule
        Temperature & 0.0 \\
        Remasking strategy & \texttt{low\_confidence} \\
        Generation length ($L$) & 256, 512 \\
        Diffusion steps & $L / 2$ \\
        \bottomrule
    \end{tabular}
    \caption{Inference hyperparameters.}
    \label{tab:eval_common}
\end{table}

\paragraph{Benchmarks.}
Table~\ref{tab:benchmarks} lists the benchmarks used for evaluation, along with the number of evaluation samples and the metric for each task.

\begin{table}[t]
    \centering
    \small
    \begin{tabular}{lcl}
        \toprule
        \textbf{Benchmark} & \textbf{Samples} & \textbf{Metric} \\
        \midrule
        GSM8K & 1{,}319 & Accuracy \\
        MATH-500 & 500 & Accuracy \\
        Countdown & 256 & Accuracy \\
        Sudoku & 256 & Accuracy \\
        HumanEval & 164 & Pass@1 \\
        \bottomrule
    \end{tabular}
    \caption{Evaluation benchmarks.}
    \label{tab:benchmarks}
\end{table}

\paragraph{Correctness Criteria.}
We describe the correctness criteria for each benchmark.
GSM8K: We extract the numerical value from the model response and check for exact match with the ground-truth integer.
MATH: We normalize LaTeX expressions and judge mathematical equivalence. For floating-point numbers, we use a tolerance of $10^{-6}$.
Countdown: We extract the arithmetic expression from the \texttt{<answer>} tag of the model response and check two conditions: (i)~the numbers used in the expression correspond one-to-one with the given numbers (each number is used exactly once), and (ii)~the evaluation result of the expression matches the target value (tolerance $10^{-5}$).
Sudoku: We extract the digit string from the \texttt{<answer>} tag and interpret it as a $4 \times 4$ board. For each empty cell (cells with value 0) in the puzzle, we compare the response with the ground-truth label cell by cell. The metric is the empty-cell accuracy (number of correctly filled cells / total number of empty cells), pooled over all problems.
HumanEval: We execute all test cases provided with the dataset and consider a response correct if and only if all test cases pass (functional correctness).

\section{Reward functions}
\label{sec:reward_functions}

For GSM8K, MATH, and Code, the reward function is a weighted sum of a correctness reward and a format reward. The format reward is an auxiliary reward that encourages responses to follow prescribed XML tags (\texttt{<reasoning>}, \texttt{<answer>}). For all tasks except coding task, we use the same reward functions as \citet{ICLR2026_2a2de7a4}. For the coding task, we design the reward function based on \citet{openr1}. We describe the reward function for each dataset below.

\paragraph{GSM8K.}
The correctness reward extracts the string within the \texttt{<answer>} tag from the model response and checks for exact string match with the ground-truth label. A match yields 2.0; a mismatch yields 0.0. The format reward consists of an XML tag count reward, a soft format reward, a strict format reward, and an integer format reward. All reward components are summed with equal weights.

\paragraph{MATH.}
The correctness reward extracts the expression within the last \texttt{\textbackslash boxed\{\}} from the model response, normalizes the LaTeX expression via regular expressions, and judges mathematical equivalence. Equivalence yields 2.0; non-equivalence yields 0.0. The format reward encourages the use of \texttt{\textbackslash boxed\{\}} and \texttt{<answer>} tags. The correctness reward and the format reward are summed with equal weights.

\paragraph{Countdown.}
We extract the arithmetic expression from the \texttt{<answer>} tag of the model response. The correctness reward is 0.0 if extraction fails, 0.1 if the expression is extracted but either the given numbers are not each used exactly once or the evaluation result does not match the target value, and 1.0 if all numbers are each used exactly once and the result matches the target.

\paragraph{Sudoku.}
We extract the digit string from the \texttt{<answer>} tag and interpret it as a $4 \times 4$ board. For each empty cell, we check for exact string match between the response and the ground-truth label. The correctness reward is the empty-cell accuracy (number of correctly filled cells / total number of empty cells), a continuous value in $[0.0, 1.0]$.

\paragraph{Code.}
We extract the Python code block from the model response and execute the test cases in a sandboxed environment. Each test case has a timeout of 5 seconds, and the correctness reward is the pass rate (number of passed tests / total number of tests). A format reward is added with a weight of 0.1. The implementation of the reward function and code evaluation follows Open-R1~\cite{openr1}.

\section{Background Supplement}
\label{app:background_details}

This section supplements the technical details omitted from the Background (Sections~\ref{subsec:mdlm_overview}--\ref{subsec:block_selection_instability}). We first describe the formulation of the forward and reverse processes of MDLM (Appendix~\ref{app:mdlm_details}), then present the general formulation of policy gradients (Appendix~\ref{app:policy_gradient}) and the intractability of $\log \pi_\theta$ in MDLM (Appendix~\ref{app:intractability}). We then provide the formal definition of the mask state in blockwise generation (Appendix~\ref{app:mask_state}), the definition of the masked-position set $\Omega_b(t_b)$ (Appendix~\ref{app:omega_def}), and the derivation of the probability that at least one block is never selected, used in Fig.~\ref{fig:block_sampling_sim_b} (Appendix~\ref{app:miss_prob}).

\subsection{Forward / Reverse Process of MDLM}
\label{app:mdlm_details}

\paragraph{Forward Process.}
The forward process of MDLM is based on absorbing-state diffusion~\citep{NEURIPS2021_958c5305} and independently masks each token. In general, a noise schedule $\alpha_t$ defines the masking probability as $1 - \alpha_t$; following \citet{NEURIPS2024_eb0b13cc}, we adopt a log-linear schedule $\alpha_t = 1 - t$. Under this schedule, for a diffusion timestep $t \in (0, 1]$, each token $o^{i}$ is replaced by the mask token $M$ with probability $t$:
\begin{align}
  q(o_{t}^{i} \mid o^{i}, t) = \begin{cases}
    1 - t & \text{if } o_{t}^{i} = o^{i}, \\
    t & \text{if } o_{t}^{i} = M,
    \notag
  \end{cases},
\end{align}
where $o_{t}^{i}$ denotes the state of position $i$ at timestep $t$.

\paragraph{Reverse Process.}
The reverse process of MDLM is realized by predicting the tokens at masked positions via $p_\theta$. In pre-training, the Evidence Lower Bound (ELBO) in~(\ref{eq:mdlm_elbo}) is maximized~\cite{NEURIPS2024_eb0b13cc}, where $\mathrm{Unif}((0, 1])$ denotes the uniform distribution over the interval $(0, 1]$.

\subsection{General Formulation of Policy Gradients}
\label{app:policy_gradient}

The gradient of the policy gradient objective $\mathcal{J}(\theta)$ in language model post-training is given by
\begin{align}
    &\nabla_{\theta}\mathcal{J}(\theta)
    \notag\\
    &=
    \mathbb{E}_{\bm{c}, \bm{o}}
    \!\left[
    R(\bm{c},\bm{o})
    \nabla_{\theta}\log \pi_{\theta}(\bm{o}\mid\bm{c})
    \right]
    \label{eq:rl1}
\end{align}
\citep{williams1992simple,schulman2017proximal}, where $\bm{c}\sim\mathcal{D}$, $\bm{o} \sim \pi_{\theta}(\cdot\mid\bm{c})$, $\mathcal{D}$ is the prompt distribution, and the reward function $R(\bm{c},\bm{o})$ is a scalar-valued function that evaluates the quality of the response $\bm{o}$ generated under the input $\bm{c}$.

\subsection{Group Relative Policy Optimization (GRPO)}
\label{app:grpo}
Following Sandwiched Policy Gradient~(SPG;~\citealt{ICLR2026_2a2de7a4}), we adopt Group Relative Policy Optimization~(GRPO;~\citealt{shao2024deepseekmath}).
Let $\{\bm{o}^j\}_{j=1}^g$ denote the $g$ responses generated by the model for an input prompt $\bm{c}$.

\paragraph{Objective for ARLM.}
Let $\mathcal{D}$ be the prompt distribution, $R(\bm{c},\bm{o})$ the reward function, and $\pi_\theta$ the generation policy with parameters $\theta$. Assuming on-policy updates and omitting clipping\footnote{GRPO originally includes importance ratio clipping as in PPO~\citep{schulman2017proximal}, but under on-policy updates the importance ratio equals 1, so clipping can be omitted.}, the GRPO objective can be written as:

\begin{align}
    &\mathcal{J}_{\text{group}}(\theta)
    \notag\\
    &=
    \mathbb{E}_{\substack{
    \bm{c},
    \bm{o}^1,\ldots, \bm{o}^g
    }}
    \!\left[
    \frac{1}{g}\sum_{j=1}^{g}
    A^j \log \pi_{\theta}(\bm{o}^j\mid\bm{c})
    \right]\!, \label{eq:grpo}
\end{align}
where $\bm{c}\sim\mathcal{D}$, $\bm{o}^1,\ldots,\bm{o}^g \overset{\mathrm{i.i.d.}}{\sim} \pi_{\mathrm{sg}[\theta]}(\cdot\mid\bm{c})$, and $A^j =R(\bm{c},\bm{o}^j) - \frac{1}{g}\sum_{k=1}^{g}R(\bm{c},\bm{o}^k)$. Here $\mathrm{sg}[\theta]$ denotes the stop-gradient operator, which fixes the parameters at sampling time so that gradients flow only through the current policy $\pi_\theta$.

\paragraph{Surrogate Objective for MDLM.}
Prior work~\citep{zhu-etal-2026-llada,NEURIPS2025_52190a03} addresses the intractability of the log-likelihood of MDLM by using the ELBO~(\ref{eq:mdlm_elbo}), a lower bound on $\log \pi_{\theta}(\bm{o}^j \mid \bm{c})$, as a surrogate objective. With this surrogate, one can optimize the following objective, which replaces $\log \pi_\theta(\bm{o}^j \mid \bm{c})$ in~(\ref{eq:grpo}) with $\mathcal{L}_{\text{ELBO}}$ in~(\ref{eq:mdlm_elbo}):
\begin{align}
    &\mathcal{J}_{\text{MDLM}}(\theta)
    \notag\\
    &=
    \mathbb{E}_{\substack{
    \bm{c},\bm{o}^1, \ldots, \bm{o}^g
    }}
    \!\left[
    \frac{1}{g}\sum_{j=1}^{g}
    A^j
    \mathcal{L}_{\text{ELBO}}(\bm{o}^j\mid\bm{c};\theta)
    \right], \notag %
\end{align}
where $\bm{c}\sim\mathcal{D}$ and $\bm{o}^1,\ldots,\bm{o}^g \overset{\mathrm{i.i.d.}}{\sim} \pi_{\mathrm{sg}[\theta]}(\cdot\mid\bm{c})$.

Furthermore, when using blockwise generation, $\mathcal{L}_{\text{ELBO}}$ in~(\ref{eq:mdlm_elbo}) is replaced by the blockwise surrogate objective $\mathcal{L}_{\text{Block}}$ in~(\ref{eq:block_obj}):
\begin{align}
    &\mathcal{J}_{\text{Block}}(\theta)
    \notag\\
    &=
    \mathbb{E}_{\substack{
    \bm{c},\bm{o}^1, \ldots, \bm{o}^g
    }}
    \!\left[
    \frac{1}{g}\sum_{j=1}^{g}
    A^j
    \mathcal{L}_{\text{Block}}(\bm{o}^j\mid\bm{c};\theta)
    \right]. \label{eq:block_grpo}
\end{align}
In practice, since $\mathcal{L}_{\text{Block}}$ is approximated with a finite number of samples, the gradients are computed by replacing $\mathcal{L}_{\text{Block}}$ in~(\ref{eq:block_grpo}) with its estimator $\hat{\mathcal{L}} \in \{\hat{\mathcal{L}}_{\text{Full}},\, \hat{\mathcal{L}}_{\text{Rand}}\}$.

\subsection{Intractability of MDLM's Log-Likelihood}
\label{app:intractability}

The likelihood of an entire sequence under the MDLM generation process requires marginalization over all possible unmasking orders $\sigma$~\cite{pmlr-v32-uria14}:
\begin{align}
\log \pi_{\theta}(\bm{o} \mid \bm{c})
= \log \sum_{\sigma \in S_{L}}
\pi_{\theta}(\bm{o}, \sigma \mid \bm{c}),
\notag %
\end{align}
where $S_{L}$ is the set of all unmasking orders for a sequence of length $L$. Since $|S_{L}|$ is on the order of $L!$, this marginalization is computationally prohibitive.

\subsection{Mask State in Blockwise Generation}
\label{app:mask_state}

We define the set of all positions masked during the generation of block $b$ as
\begin{align}
\mathcal{S}_b(\mathcal{M}_{t_b})
&=
\mathcal{M}_{t_b}
\cup
\bigcup_{b'>b}\mathcal{B}_{b'}.
\label{eq:extended_mask_set}
\end{align}
This set includes the masked positions within block $b$ as well as all positions of subsequent blocks. The mask state of each position during the generation of block $b$ is then given by
\begin{align}
    o_{b,\mathcal{M}_{t_b}}^{i}
    =
    \begin{cases}
    o^{i}
    & i \notin \mathcal{S}_b(\mathcal{M}_{t_b}),
    \\[1mm]
    M
    & i \in \mathcal{S}_b(\mathcal{M}_{t_b}).
    \end{cases}
    \label{eq:mask_state}
\end{align}
That is, the tokens in blocks preceding $b$ and the unmasked tokens within block $b$ are retained, while all positions in $\mathcal{S}_b$ are masked.

\subsection{Formal Definition of \texorpdfstring{$\Omega_b(t_b)$}{}}
\label{app:omega_def}

We define the set of all possible masked-position subsets within block $b$ under masking ratio $t_b$ as
\begin{align}
    \Omega_b(t_b)
    &:=
    \bigl\{
    \mathcal{M}\subseteq \mathcal{B}_b
    \bigm|
    |\mathcal{M}|=\max(1,\lfloor t_b|\mathcal{B}_b|\rfloor)
    \bigr\}.
    \notag
\end{align}
When $L$ is divisible by $B$, we have $|\mathcal{B}_b| = B$, and $\Omega_b(t_b)$ is the set of all $\max(1, \lfloor t_b B \rfloor)$-element subsets of $\mathcal{B}_b$.

\subsection{Probability That at Least One Block Is Never Selected}
\label{app:miss_prob}

We derive the probability, used in Fig.~\ref{fig:block_sampling_sim_b}, that at least one block is never selected when sampling $K$ times i.i.d.\ uniformly from $C$ blocks.

For each block $b \in \{0, \dots, C - 1\}$, let $A_b$ denote the event that block $b$ is never selected in $K$ samples. The desired probability is $\Pr\!\left(\bigcup_{b=0}^{C-1} A_b\right)$.

For each $A_b$, $\Pr(A_b) = \left(\frac{C - 1}{C}\right)^K$, and for any set of $s$ blocks $\{b_1, \dots, b_s\}$,
\begin{align}
    \Pr\!\left(\bigcap_{m=1}^{s} A_{b_m}\right)
    = \left(\frac{C - s}{C}\right)^K.
    \notag
\end{align}
Applying the inclusion-exclusion principle,
\begin{align}
    &\Pr\!\left(\bigcup_{b=0}^{C-1} A_b\right)
    \notag \\
    &= \sum_{s=1}^{C} (-1)^{s+1} \binom{C}{s} \left(\frac{C - s}{C}\right)^K.
    \label{eq:miss_prob}
\end{align}
For example, when $K = C = 4$, (\ref{eq:miss_prob}) gives $\Pr \approx 0.91$, which is consistent with Fig.~\ref{fig:block_sampling_sim_b}.

\section{Variance Analysis for \texorpdfstring{$K \neq C$}{}}
\label{app:variance_kneqc}

The estimator $\hat{\mathcal{L}}_{\text{Full}}$ in (\ref{eq:full_block_estimator}) handles only $K = C$.
We denote its extension to arbitrary $K \geq 1$ by $\hat{\mathcal{L}}_{\text{Full}}^{(K)}$ and derive the variance ratio relative to $\hat{\mathcal{L}}_{\text{Rand}}$.
Following the notation of Section~\ref{sec:variance_reduction}, we additionally define
\begin{align}
    \lambda
:=
    \frac{\sigma_{\mathrm{between}}^2}{\sigma_{\mathrm{between}}^2 + \bar{\sigma}_{\mathrm{within}}^2}
    \in [0, 1]
    \label{eq:lambda_def}
\end{align}
to denote the contribution of the between-group variance to the total variance.

\subsection{Definition of \texorpdfstring{$\hat{\mathcal{L}}_{\text{Full}}^{(K)}$}{}}
\label{app:full_K_def}

Decompose the sample size $K$ as $K = qC + r$, where $q:= \lfloor K/C \rfloor \geq 0$ and $r:= K \bmod C \in \{0, \dots, C-1\}$, and define the estimator as the sum of a deterministic part and a random part:
\begin{align}
    &\hat{\mathcal{L}}_{\text{Full}}^{(K)}(\bm{o}\mid\bm{c};\theta)
    \notag\\
    &:=
    \frac{1}{K}
    \!\left[
    \sum_{b=0}^{C-1}\sum_{j=1}^{q}
    \ell_b^{(j)}
    +
    \sum_{k=1}^{r}
    \ell_{b_k}
    \right].
    \label{eq:full_block_general}
\end{align}
In the first term (the deterministic part), for each block $b$ we sample $q$ pairs $(t_b^{(j)}, \mathcal{M}_{t_b}^{(j)})$ independently and i.i.d., and write the corresponding loss as $\ell_b^{(j)}:= \ell_{b, t_b^{(j)}, \mathcal{M}_{t_b}^{(j)}}$.
In the second term (the random part), we draw $r$ block indices $\{b_1, \dots, b_r\}$ uniformly \textbf{without replacement} from $\{0, \dots, C-1\}$.
For each $b_k$, we sample $(t_{b_k}, \mathcal{M}_{t_{b_k}})$ independently and i.i.d., and compute the loss $\ell_{b_k}:= \ell_{b_k, t_{b_k}, \mathcal{M}_{t_{b_k}}}$.
The total number of samples is $qC + r = K$ (only the deterministic part if $r = 0$; only the $K$ without-replacement samples if $q = 0$).

\subsection{Derivation of the Variance}
\label{app:full_K_derivation}

The deterministic part $S_\text{det}:= \sum_{b,j} \ell_b^{(j)}$ and the random part $S_\text{rand}:= \sum_{k} \ell_{b_k}$ are independent because they use independent $(t, M)$ sample sequences. We derive each variance.

\paragraph{Variance of the deterministic part.}
The $qC$ samples $\ell_b^{(j)}$ are all independent, and using the relation $\sum_{b=0}^{C-1} \sigma_b^2 = C \bar{\sigma}_{\mathrm{within}}^2$ from (\ref{eq:sig_within}), we obtain
\begin{align}
    \mathrm{Var}[S_\text{det}]
    = \sum_{b=0}^{C-1} q \sigma_b^2
    = qC \bar{\sigma}_{\mathrm{within}}^2.
    \label{eq:var_S_det}
\end{align}

\paragraph{Variance of the random part.}
Conditioned on $\mathbf{b} = (b_1, \dots, b_r)$, each $X_k:= \ell_{b_k}$ is independent. Applying the law of total variance to $\sum_k X_k$,
\begin{align}
    \mathrm{Var}[S_\text{rand}]
    &=
    \mathbb{E}\!\left[\sum_{k=1}^{r} \sigma_{b_k}^2 \mid \mathbf{b}\right]
    +
    \mathrm{Var}\!\left[\sum_{k=1}^{r} \mu_{b_k}\right] \notag\\
    &=
    r \bar{\sigma}_{\mathrm{within}}^2
    +
    \mathrm{Var}\!\left[\sum_{k=1}^{r} \mu_{b_k}\right].
    \label{eq:var_S_rand_step1}
\end{align}
For the second term, using $\mathrm{Var}[\mu_{b_k}] = \sigma_{\mathrm{between}}^2$ and the without-replacement covariance $\mathrm{Cov}[\mu_{b_k}, \mu_{b_{k'}}] = -\sigma_{\mathrm{between}}^2 / (C - 1)$ for $k \neq k'$, we have
\begin{align}
    \mathrm{Var}\!\left[\sum_{k=1}^{r} \mu_{b_k}\right]
    &= r \sigma_{\mathrm{between}}^2 - \frac{r(r-1)}{C-1} \sigma_{\mathrm{between}}^2 \notag\\
    &= \frac{r(C-r)}{C-1} \sigma_{\mathrm{between}}^2,
    \label{eq:var_mean_blocks}
\end{align}
whence
\begin{align}
    \mathrm{Var}[S_\text{rand}]
    = r \bar{\sigma}_{\mathrm{within}}^2
    + \frac{r(C-r)}{C-1} \sigma_{\mathrm{between}}^2.
    \label{eq:var_S_rand}
\end{align}

\paragraph{Total variance.}
Since $S_\text{det} \perp S_\text{rand}$, combining (\ref{eq:var_S_det}) and (\ref{eq:var_S_rand}) gives
\begin{align}
    \mathrm{Var}\!\left[\hat{\mathcal{L}}_{\text{Full}}^{(K)}\right]
    &= \frac{1}{K^2}\!\left(\mathrm{Var}[S_\text{det}] + \mathrm{Var}[S_\text{rand}]\right) \notag\\
    &= \frac{\bar{\sigma}_{\mathrm{within}}^2}{K}
    + \frac{r(C-r)}{K^2(C-1)} \sigma_{\mathrm{between}}^2.
    \label{eq:var_full_general}
\end{align}
Together with (\ref{eq:var_rand}), the variance ratio can be written, with $r:= K \bmod C \in \{0, \dots, C-1\}$, as
\begin{align}
    \frac{\mathrm{Var}\!\left[\hat{\mathcal{L}}_{\text{Full}}^{(K)}\right]}{\mathrm{Var}\!\left[\hat{\mathcal{L}}_{\text{Rand}}\right]}
    &= 1 - \lambda \cdot \frac{K(C-1) - r(C-r)}{K(C-1)} \notag\\
    &= 1 - \lambda + \lambda \cdot \frac{r(C-r)}{K(C-1)}
    \label{eq:var_ratio_unified}
\end{align}
The second term on the right-hand side is a nonnegative correction term that vanishes only when $r = 0$.
The variance ratio is therefore at most $1$ for all $K \geq 1$.

\subsection{Reduction to Special Cases}
\label{app:full_K_corollaries}

Expression (\ref{eq:var_ratio_unified}) immediately yields the following special cases, summarized in Table~\ref{tab:variance_ratio_kneqc}.

\begin{table}[t]
    \small
    \centering
    \setlength{\tabcolsep}{4pt}
    \begin{tabular}{lc}
        \toprule
        Setting & Variance ratio vs $\hat{\mathcal{L}}_{\text{Rand}}$ \\
        \midrule
        General ($K = qC + r$) & $1 - \lambda + \lambda \cdot \dfrac{r(C-r)}{K(C-1)}$ \\
        \midrule
        $K = 1$ & $1$ \\
        $K < C$ ($q = 0$, $r = K$) & $1 - \lambda \cdot \dfrac{K-1}{C-1}$ \\
        $K = mC$ ($m \geq 1$, $r = 0$) & $1 - \lambda$ \\
        \bottomrule
    \end{tabular}
    \caption{Variance ratio of $\hat{\mathcal{L}}_{\text{Full}}^{(K)}$ relative to $\hat{\mathcal{L}}_{\text{Rand}}$. Here $K = qC + r$ with $q = \lfloor K/C \rfloor$ and $r = K \bmod C \in \{0, \dots, C-1\}$, and $\lambda$ is defined in (\ref{eq:lambda_def}). The case $K = C$ corresponds to $K = mC$ with $m = 1$.}
    \label{tab:variance_ratio_kneqc}
\end{table}

\paragraph{\texorpdfstring{$K = mC$ ($m \geq 1$, $r = 0$)}{}.}
Since $r(C-r) = 0$, the variance ratio equals $1 - \lambda$.
This is a constant independent of $K$ and $m$.
Hence even as $K \to \infty$, the variance ratio remains at the same value as in the case $K = C$.

\paragraph{\texorpdfstring{$K = C$ ($q = 1$, $r = 0$)}{}.}
This is the special case of $K = mC$ with $m = 1$.
The variance ratio equals $1 - \lambda$, consistent with evaluating (\ref{eq:var_ratio}) in the main text at $K = C$.

\paragraph{\texorpdfstring{$K < C$ ($q = 0$, $r = K$)}{}.}
From $K(C-1) - K(C-K) = K(K-1)$, the variance ratio equals $1 - \lambda \cdot (K-1)/(C-1)$.
It equals $1$ at $K = 1$ (matching $\hat{\mathcal{L}}_{\text{Rand}}$) and decreases monotonically in $K$.

\subsection{Behavior of the Correction Term and the Variance Ratio as a Function of \texorpdfstring{$K$}{}}
\label{app:full_K_nonmonotone}

The correction term $\lambda \cdot r(C-r)/(K(C-1))$ in (\ref{eq:var_ratio_unified}) vanishes only when $r = 0$, so the variance-ratio gap between our method and $\hat{\mathcal{L}}_{\text{Rand}}$ is maximized when $K$ is a multiple of $C$ (i.e., $K = mC$).
When $K$ deviates from a multiple of $C$ ($r \neq 0$), the correction term becomes positive and the gap shrinks; for example, with $C = 4$ and $\lambda = 0.90$, the variance ratio is $\approx 0.10$ at $K = 4$ but $\approx 0.28$ at $K = 5$. Here $\lambda \approx 0.90$ is a representative value consistent with the empirical measurements in Table~\ref{tab:variance_ratio} (variance ratio $1 - \lambda \approx 0.08$--$0.12$).
This occurs because the single random sample added at $K = C+1$ reintroduces stochastic variation in the block index.
Nevertheless, the variance ratio remains at most $1$ for all $K \geq 1$.

\section{Empirical Verification for \texorpdfstring{$K \neq C$}{}}
\label{app:k_lt_c_empirical}

The special cases derived in Appendix~\ref{app:full_K_corollaries} describe the behavior of the variance ratio between $\hat{\mathcal{L}}_{\text{Full}}^{(K)}$ and $\hat{\mathcal{L}}_{\text{Rand}}$ for $K \leq C$ and for multiples of $C$: the ratio equals $1$ at $K = 1$ (the two estimators coincide), decreases monotonically to $1 - \lambda$ at $K = C$, and remains at the constant value $1 - \lambda$ independently of $K$ and $m$ for $K = mC$ with $m \geq 2$.
This section reports training experiments on both sides of the boundary $K = C$ to examine whether the advantage of our method depends on the specific setting $K = C = 4$.

\subsection{Left of the Boundary: Empirical Evidence for \texorpdfstring{$K \leq C$}{}}
\label{app:ext_kleqc_evidence}

The special case $K \leq C$ ($q = 0$, $r = K$) predicts that the variance ratio decreases monotonically from $1$ at $K = 1$ to $1 - \lambda$ at $K = C$.
This subsection reports a training experiment at the point $K = 2$, $C = 4$.

We newly train the baselines (SPG and $\hat{\mathcal{L}}_{\text{Rand}}$) with $K = 2$ and the number of training iterations $I = 8{,}000$, and reuse the result of our method at $K = C = 4$, $I = 4{,}000$ from Table~\ref{tab:main_results_styled}. The baselines perform 4{,}000 updates, as in the main experiment, and in each update they compute two independent $K = 2$ estimates and sum the resulting gradients. Thus, $I = 8{,}000$ is the total number of these estimates.
The total number of block evaluations $K \cdot I = 16{,}000$ is matched across the two settings\footnote{In general, for the proposed method with iteration count $I_\text{Full}$ and the random block sampling baseline with iteration count $I_\text{Rand}$, equalizing the total number of block evaluations corresponds to $C \cdot I_\text{Full} = K \cdot I_\text{Rand}$, i.e., $I_\text{Rand} = (C / K) \cdot I_\text{Full}$. For example, if $C = 8, K = 2$, then $I_\text{Rand} = 4 \cdot I_\text{Full}$.}, to examine whether the baselines catch up with our method when $K$ is halved and the number of iterations is doubled.
All other hyperparameters are identical to those in Section~\ref{sec:main-exp-setup} and Appendix~\ref{app:training_details}; only the baselines' $K$ and $I$ are changed.

\begin{table*}[t]
    \small
    \centering
    \setlength{\tabcolsep}{3pt}
    \begin{tabular}{l cccccc}
        \toprule
        Method & GSM8K & MATH-500 & Countdown & Sudoku & HumanEval & Avg. \\
        \midrule
        SPG & 81.88 & 35.20 & 64.84 & 89.76 & 30.49 & 60.43 \\
        \rowcolor{okabeorange!8}
        $\hat{\mathcal{L}}_{\mathrm{Rand}}$ & 81.58 & 33.40 & 54.30 & 80.96 & 31.10 & 56.27 \\
        \rowcolor{okabeblue!8}
        $\hat{\mathcal{L}}_{\mathrm{Full}}$ (ours) & \textbf{82.26} & \textbf{38.00} & \textbf{66.41} & \textbf{91.39} & \textbf{36.59} & \textbf{62.93} \\
        \bottomrule
    \end{tabular}
    \caption{Empirical verification for $K < C$. Baselines (SPG and $\hat{\mathcal{L}}_{\mathrm{Rand}}$) are re-trained with $K = 2$, $C = 4$, $I = 8{,}000$, and our method $\hat{\mathcal{L}}_{\mathrm{Full}}$ uses $K = C = 4$, $I = 4{,}000$ (reused from Table~\ref{tab:main_results_styled}). Both settings have the same total number of block evaluations $K \cdot I = 16{,}000$.}
    \label{tab:k_lt_c_results}
\end{table*}

As shown in Table~\ref{tab:k_lt_c_results}, $\hat{\mathcal{L}}_{\text{Full}}$ (avg.\ 62.93) outperforms both $\hat{\mathcal{L}}_{\text{Rand}}$ (56.27) and SPG (60.43) on all five benchmarks, and the advantage is preserved under a matched number of block evaluations.\footnote{Because the training settings differ from those of Table~\ref{tab:main_results_styled}, the baseline values in Table~\ref{tab:k_lt_c_results} do not match those in Table~\ref{tab:main_results_styled}.}

\subsection{\texorpdfstring{Right of the Boundary: Empirical Evidence}{Right of the Boundary: Empirical Evidence} for \texorpdfstring{$K > C$}{} (Oversampling)}
\label{app:ext_kgtc_oversample}

The setting $K > C$ increases the number of block evaluations per iteration beyond $C$ and thus incurs higher compute cost, so it is not normally used. However, this subsection also examines the $K > C$ regime.
The $K = mC$ special case states that the variance ratio is fixed at $1 - \lambda$ independently of $K$ and $m$.
We examine two points with $C = 4$ fixed: a point without the correction term, $K = mC$ ($K = 8$, $m = 2$; the $K = mC$ special case predicts that the variance ratio returns to $1 - \lambda$, the same as at $K = C$), and a point with a positive correction term predicted by Appendix~\ref{app:full_K_nonmonotone} ($K = 6$, $r = 2$; the variance ratio is larger than $1 - \lambda$ but still strictly smaller than $1$).

\begin{table*}[t]
    \small
    \centering
    \setlength{\tabcolsep}{3pt}
    \begin{tabular}{l cc cc cc}
        \toprule
        & \multicolumn{2}{c}{$K = 4$} & \multicolumn{2}{c}{$K = 6$} & \multicolumn{2}{c}{$K = 8$} \\
        \cmidrule(lr){2-3}\cmidrule(lr){4-5}\cmidrule(lr){6-7}
        Method & GSM8K & MATH-500 & GSM8K & MATH-500 & GSM8K & MATH-500 \\
        \midrule
        \rowcolor{okabeorange!8}
        $\hat{\mathcal{L}}_{\mathrm{Rand}}$ & 76.20 & 34.40 & 80.89 & 34.00 & 80.97 & 34.00 \\
        \rowcolor{okabeblue!8}
        $\hat{\mathcal{L}}_{\mathrm{Full}}^{(K)}$ (ours) & \textbf{82.26} & \textbf{38.00} & 80.89 & \textbf{36.20} & \textbf{81.43} & \textbf{34.20} \\
        \bottomrule
    \end{tabular}
    \caption{Empirical verification for $K > C$ ($C = 4$ fixed, $I = 4{,}000$). The $K = 4$ column is reused from Table~\ref{tab:main_results_styled}. The $K = 6$ and $K = 8$ columns are new experiments.}
    \label{tab:k_gt_c_results}
\end{table*}

Table~\ref{tab:k_gt_c_results} reports the accuracy comparison between $\hat{\mathcal{L}}_{\text{Full}}^{(K)}$ and $\hat{\mathcal{L}}_{\text{Rand}}$ on GSM8K and MATH-500 at the two points.
At $K = 8$, the variance ratio between $\hat{\mathcal{L}}_{\text{Full}}^{(K)}$ and $\hat{\mathcal{L}}_{\text{Rand}}$ remains at $1 - \lambda$ as predicted by the $K = mC$ theory, while the empirical accuracy gap in Table~\ref{tab:k_gt_c_results} is smaller than at $K = 4$ (GSM8K $+6.06 \to +0.46$; MATH-500 $+3.60 \to +0.20$). At $K = 6$, the accuracy gap on GSM8K is smaller than at $K = 8$. On GSM8K specifically, both methods reach the same accuracy of $80.89$ at $K = 6$; here, $\hat{\mathcal{L}}_{\text{Rand}}$ approaches the GSM8K performance level attained by $\hat{\mathcal{L}}_{\text{Full}}^{(K)}$ across this $K$ range (within a $\pm 1$-point band around $81\%$). MATH-500, by contrast, still shows a $+2.20$-point advantage for $\hat{\mathcal{L}}_{\text{Full}}^{(K)}$ at $K = 6$.
In summary, within the settings we examined, the gain of our method is largest in the low-sample regime $K = C$ and generally shrinks as the sample size $K$ grows beyond the number of blocks $C$.
\section{Generalization to a Doubled Block Count \texorpdfstring{$C$}{}}
\label{app:c_scalability}

The variance-reduction derivation in Section~\ref{sec:variance_reduction} treats $C$ as a general parameter and does not depend on its specific value.
In this section, we retrain $\hat{\mathcal{L}}_{\text{Full}}$ and $\hat{\mathcal{L}}_{\text{Rand}}$ on GSM8K and MATH-500 with $C = K = 8$ and compare the results with the main experiment ($C = K = 4$) in Table~\ref{tab:c8_results}, showing that the advantage of our method is preserved when $C$ is doubled.
Code, Countdown, and Sudoku are excluded from this verification due to limited computation time and resources.

\begin{table*}[t]
    \small
    \centering
    \setlength{\tabcolsep}{4pt}
    \begin{tabular}{l cc cc}
        \toprule
        & \multicolumn{2}{c}{$C = K = 4$} & \multicolumn{2}{c}{$C = K = 8$} \\
        \cmidrule(lr){2-3}\cmidrule(lr){4-5}
        Method & GSM8K & MATH-500 & GSM8K & MATH-500 \\
        \midrule
        \rowcolor{okabeorange!8}
        $\hat{\mathcal{L}}_{\mathrm{Rand}}$ & 76.20 & 34.40 & 80.52 & 31.80 \\
        \rowcolor{okabeblue!8}
        $\hat{\mathcal{L}}_{\mathrm{Full}}$ (ours) & \textbf{82.26} & \textbf{38.00} & \textbf{81.35} & \textbf{36.80} \\
        \bottomrule
    \end{tabular}
    \caption{Empirical verification of doubling $C$. Comparison between the main experiment ($C = K = 4$) and the re-trained models with $C = K = 8$. The $C = K = 4$ columns are reused from Table~\ref{tab:main_results_styled}. All other hyperparameters are identical to those of the main experiment (LLaDA-8B-Instruct, $I = 4{,}000$). Code, Countdown, and Sudoku are excluded due to limited computation time and resources. \textbf{Bold} indicates the higher value within each $C = K$ setting.}
    \label{tab:c8_results}
\end{table*}

As shown in Table~\ref{tab:c8_results}, $\hat{\mathcal{L}}_{\text{Full}}$ outperforms $\hat{\mathcal{L}}_{\text{Rand}}$ on both GSM8K and MATH-500 at $C = K = 8$ as well, showing that the advantage of our method is preserved when $C$ is doubled.

\section{Detailed Estimation Quality Analysis}
\label{sec:detailed_estimation_analysis}

This section extends the estimation quality analysis conducted on GSM8K in Section~\ref{sec:analysis} to all five datasets (GSM8K, MATH-500, Countdown, Sudoku, and Code). The experimental setup is identical to Section~\ref{sec:analysis}. We first report the overall statistics of estimation errors (mean, standard deviation, and median) and compare per-sample variances in tables (Appendix~\ref{subsec:error_stats}). We then present five types of analysis results for each dataset in figures (Appendix~\ref{subsec:per_dataset_details}): Spearman rank correlation $\rho$ and RMSE with respect to the reference value (estimation quality), tail probability $\Pr(|\hat{\mathcal{L}} - \mathcal{L}_{\mathrm{ref}}| > \tau)$ (frequency of large estimation errors), the density distribution of estimation errors, and scatter plots of per-sample variances.

\subsection{Estimation Error Statistics}
\label{subsec:error_stats}

Table~\ref{tab:error_stats} shows the statistics of the estimation error $|\hat{\mathcal{L}} - \mathcal{L}_{\mathrm{ref}}|$ across all five datasets. Across all datasets, $\hat{\mathcal{L}}_{\text{Full}}$ achieves substantially lower mean error and median compared with $\hat{\mathcal{L}}_{\text{Rand}}$. The improvement is particularly notable on Sudoku, where the mean error is reduced approximately fourfold from $105.95$ to $26.42$. On Code, the absolute errors are larger than on the other datasets, which is likely due to the longer generation length ($L = 512$); nevertheless, $\hat{\mathcal{L}}_{\text{Full}}$ improves the mean error of $\hat{\mathcal{L}}_{\text{Rand}}$ by approximately threefold.

Table~\ref{tab:per_sample_variance} compares the per-sample variances of the absolute error. The proportion of samples for which this variance is lower for $\hat{\mathcal{L}}_{\text{Full}}$ than for $\hat{\mathcal{L}}_{\text{Rand}}$ is at least $98.9\%$ across all datasets, reaching $100.0\%$ on Sudoku. The ratio of the mean variances ranges from $7.35\times$ to $11.28\times$, and all differences are statistically significant under the Wilcoxon signed-rank test with $p < 10^{-100}$.

\begin{table}[t]
    \centering
    \small
    \begin{tabular}{llrrr}
        \toprule
        & & \textbf{Mean} & \textbf{Std} & \textbf{Median} \\
        \midrule
        \multirow{2}{*}{GSM8K}
        & $\hat{\mathcal{L}}_{\text{Rand}}$
        & 109.32 & 85.14 & 90.81 \\
        & $\hat{\mathcal{L}}_{\text{Full}}$
        & 36.25 & 31.82 & 28.85 \\
        \midrule
        \multirow{2}{*}{MATH}
        & $\hat{\mathcal{L}}_{\text{Rand}}$
        & 118.14 & 90.89 & 99.45 \\
        & $\hat{\mathcal{L}}_{\text{Full}}$
        & 36.64 & 33.32 & 29.18 \\
        \midrule
        \multirow{2}{*}{Countdown}
        & $\hat{\mathcal{L}}_{\text{Rand}}$
        & 131.66 & 98.14 & 113.46 \\
        & $\hat{\mathcal{L}}_{\text{Full}}$
        & 37.82 & 32.09 & 30.90 \\
        \midrule
        \multirow{2}{*}{Sudoku}
        & $\hat{\mathcal{L}}_{\text{Rand}}$
        & 105.95 & 80.04 & 92.64 \\
        & $\hat{\mathcal{L}}_{\text{Full}}$
        & 26.42 & 23.79 & 20.91 \\
        \midrule
        \multirow{2}{*}{Code}
        & $\hat{\mathcal{L}}_{\text{Rand}}$
        & 297.49 & 233.67 & 247.40 \\
        & $\hat{\mathcal{L}}_{\text{Full}}$
        & 99.69 & 86.04 & 80.44 \\
        \bottomrule
    \end{tabular}
    \caption{Statistics of the estimation error
    $|\hat{\mathcal{L}} - \mathcal{L}_{\text{ref}}|$
    ($K = 4$, 50 independent repetitions).
    The reference value is computed with $K = 1024$.}
    \label{tab:error_stats}
\end{table}

\begin{table}[t]
    \centering
    \small
    \begin{tabular}{lrr}
        \toprule
        & \textbf{Full $<$ Rand (\%)} & \textbf{Var.\ ratio} \\
        \midrule
        GSM8K & 99.5 & $7.35\times$ \\
        MATH & 98.9 & $7.47\times$ \\
        Countdown & 99.7 & $9.71\times$ \\
        Sudoku & 100.0 & $11.28\times$ \\
        Code & 99.6 & $7.36\times$ \\
        \bottomrule
    \end{tabular}
    \caption{Comparison of per-sample variances of the absolute error $|\hat{\mathcal{L}} - \mathcal{L}_{\mathrm{ref}}|$ ($K = 4$, computed from 50 independent repetitions).
    \textbf{Full $<$ Rand} denotes the proportion of samples for which $\hat{\mathcal{L}}_{\text{Full}}$ has lower variance than $\hat{\mathcal{L}}_{\text{Rand}}$.
    \textbf{Var.\ ratio} denotes the ratio of the mean variances ($\hat{\mathcal{L}}_{\text{Rand}} / \hat{\mathcal{L}}_{\text{Full}}$).
    Wilcoxon signed-rank test $p < 10^{-100}$ for all datasets.}
    \label{tab:per_sample_variance}
\end{table}

\begin{figure*}[t]
    \centering
    \begin{minipage}[c]{0.03\linewidth}~\end{minipage}\hfill
    \begin{minipage}[c]{0.18\linewidth}\centering Spearman $\rho$\end{minipage}\hfill
    \begin{minipage}[c]{0.18\linewidth}\centering RMSE\end{minipage}\hfill
    \begin{minipage}[c]{0.18\linewidth}\centering Error tail\end{minipage}\hfill
    \begin{minipage}[c]{0.18\linewidth}\centering Error dist.\ (a)\end{minipage}\hfill
    \begin{minipage}[c]{0.18\linewidth}\centering Error dist.\ (b)\end{minipage}
    \par\vspace{2pt}
    \begin{minipage}[c]{0.03\linewidth}\centering\rotatebox{90}{GSM8K}\end{minipage}\hfill
    \begin{minipage}[c]{0.18\linewidth}\centering\includegraphics[width=\linewidth]{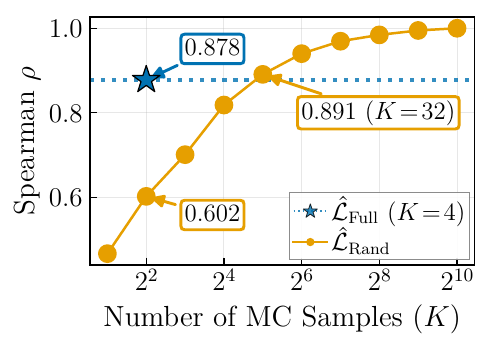}\end{minipage}\hfill
    \begin{minipage}[c]{0.18\linewidth}\centering\includegraphics[width=\linewidth]{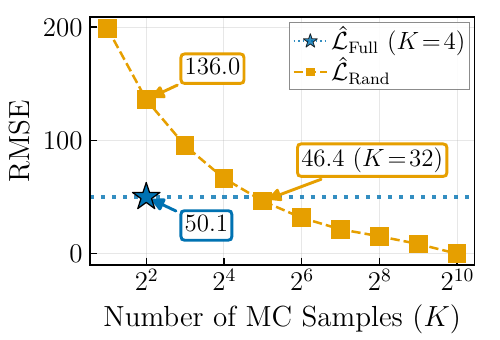}\end{minipage}\hfill
    \begin{minipage}[c]{0.18\linewidth}\centering\includegraphics[width=\linewidth]{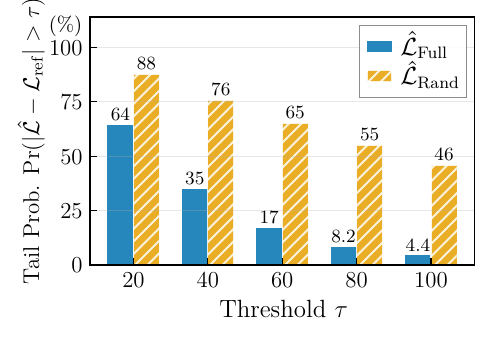}\end{minipage}\hfill
    \begin{minipage}[c]{0.18\linewidth}\centering\includegraphics[width=\linewidth]{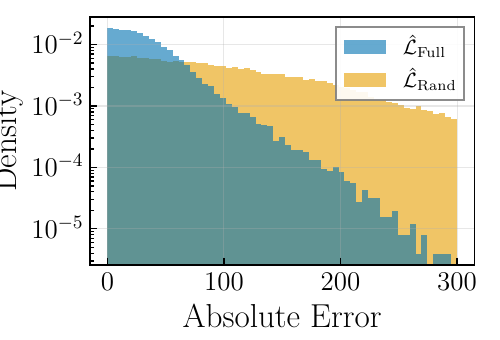}\end{minipage}\hfill
    \begin{minipage}[c]{0.18\linewidth}\centering\includegraphics[width=\linewidth]{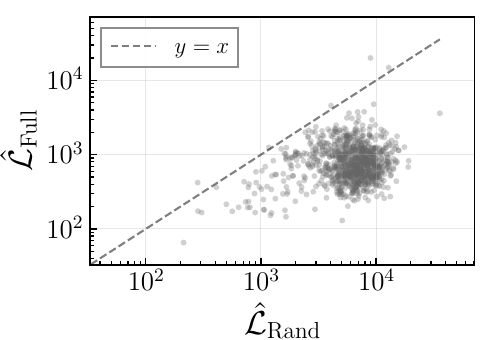}\end{minipage}
    \par\vspace{4pt}
    \begin{minipage}[c]{0.03\linewidth}\centering\rotatebox{90}{MATH-500}\end{minipage}\hfill
    \begin{minipage}[c]{0.18\linewidth}\centering\includegraphics[width=\linewidth]{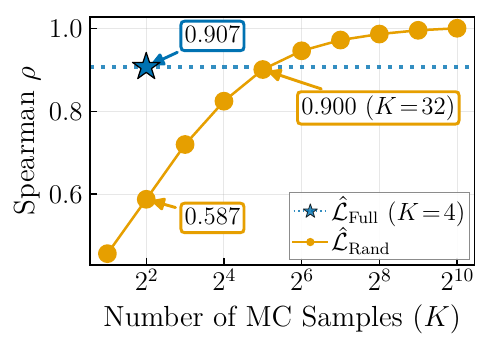}\end{minipage}\hfill
    \begin{minipage}[c]{0.18\linewidth}\centering\includegraphics[width=\linewidth]{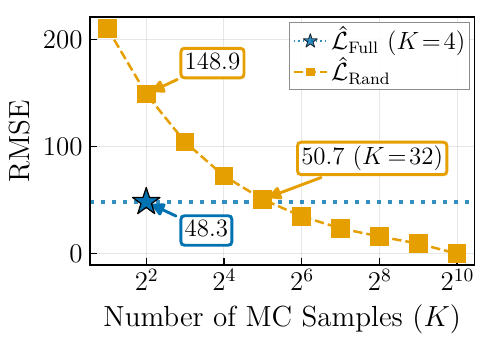}\end{minipage}\hfill
    \begin{minipage}[c]{0.18\linewidth}\centering\includegraphics[width=\linewidth]{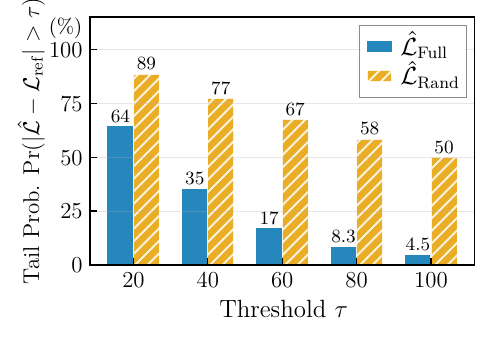}\end{minipage}\hfill
    \begin{minipage}[c]{0.18\linewidth}\centering\includegraphics[width=\linewidth]{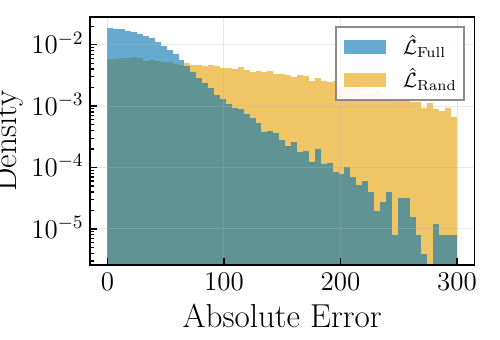}\end{minipage}\hfill
    \begin{minipage}[c]{0.18\linewidth}\centering\includegraphics[width=\linewidth]{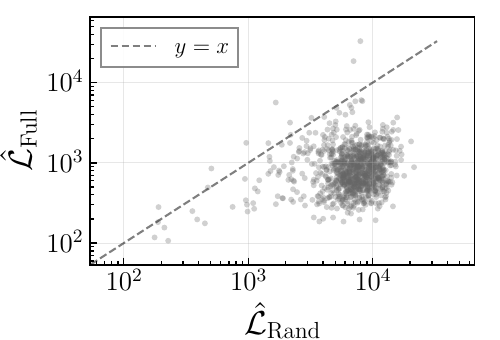}\end{minipage}
    \par\vspace{4pt}
    \begin{minipage}[c]{0.03\linewidth}\centering\rotatebox{90}{Countdown}\end{minipage}\hfill
    \begin{minipage}[c]{0.18\linewidth}\centering\includegraphics[width=\linewidth]{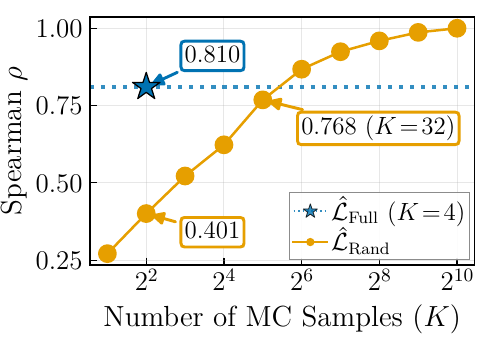}\end{minipage}\hfill
    \begin{minipage}[c]{0.18\linewidth}\centering\includegraphics[width=\linewidth]{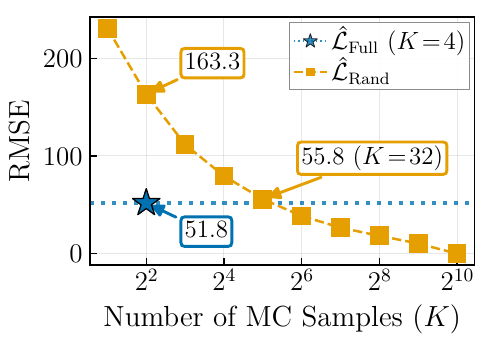}\end{minipage}\hfill
    \begin{minipage}[c]{0.18\linewidth}\centering\includegraphics[width=\linewidth]{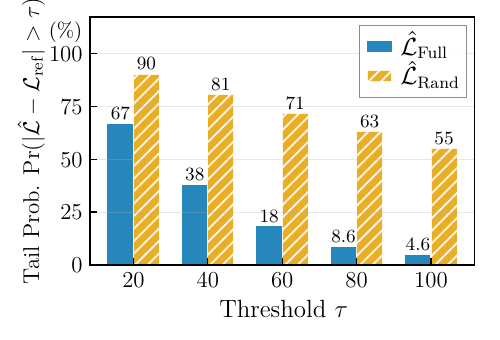}\end{minipage}\hfill
    \begin{minipage}[c]{0.18\linewidth}\centering\includegraphics[width=\linewidth]{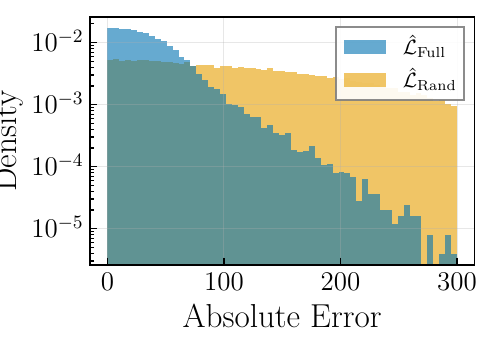}\end{minipage}\hfill
    \begin{minipage}[c]{0.18\linewidth}\centering\includegraphics[width=\linewidth]{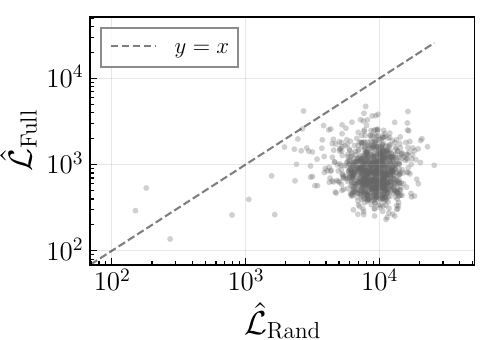}\end{minipage}
    \par\vspace{4pt}
    \begin{minipage}[c]{0.03\linewidth}\centering\rotatebox{90}{Sudoku}\end{minipage}\hfill
    \begin{minipage}[c]{0.18\linewidth}\centering\includegraphics[width=\linewidth]{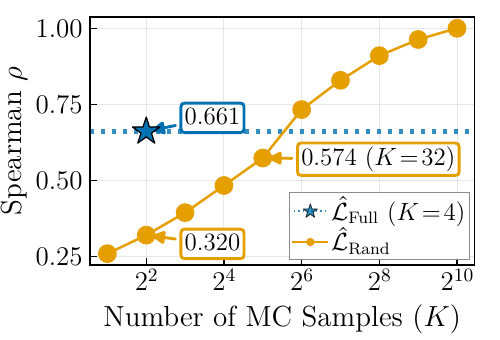}\end{minipage}\hfill
    \begin{minipage}[c]{0.18\linewidth}\centering\includegraphics[width=\linewidth]{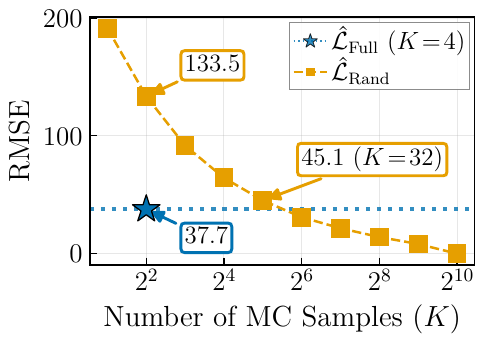}\end{minipage}\hfill
    \begin{minipage}[c]{0.18\linewidth}\centering\includegraphics[width=\linewidth]{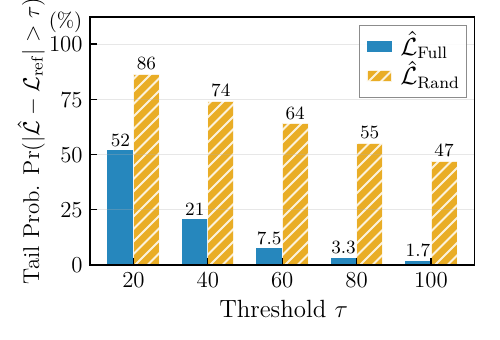}\end{minipage}\hfill
    \begin{minipage}[c]{0.18\linewidth}\centering\includegraphics[width=\linewidth]{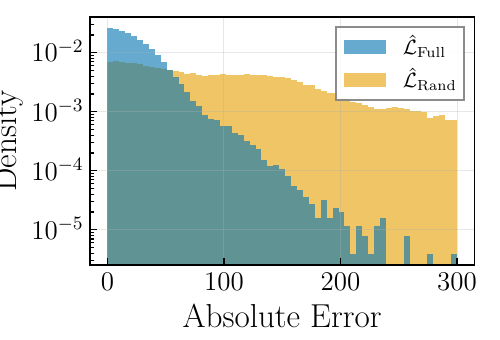}\end{minipage}\hfill
    \begin{minipage}[c]{0.18\linewidth}\centering\includegraphics[width=\linewidth]{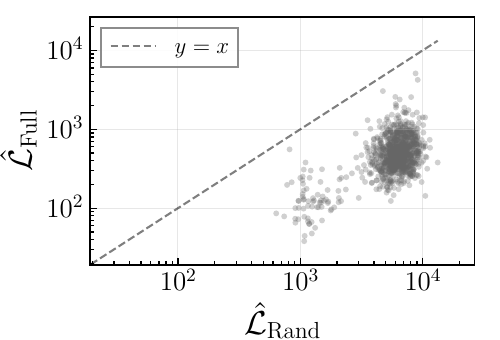}\end{minipage}
    \par\vspace{4pt}
    \begin{minipage}[c]{0.03\linewidth}\centering\rotatebox{90}{Code}\end{minipage}\hfill
    \begin{minipage}[c]{0.18\linewidth}\centering\includegraphics[width=\linewidth]{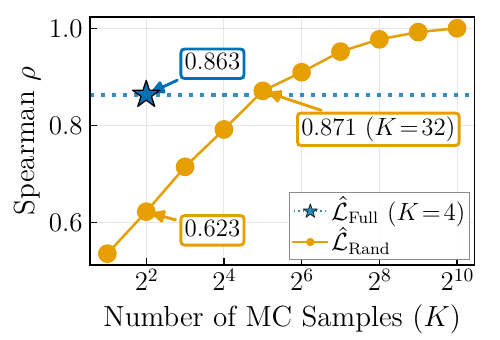}\end{minipage}\hfill
    \begin{minipage}[c]{0.18\linewidth}\centering\includegraphics[width=\linewidth]{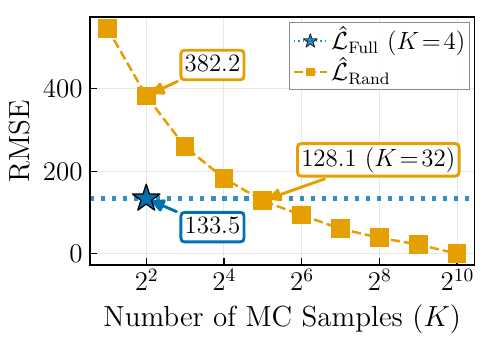}\end{minipage}\hfill
    \begin{minipage}[c]{0.18\linewidth}\centering\includegraphics[width=\linewidth]{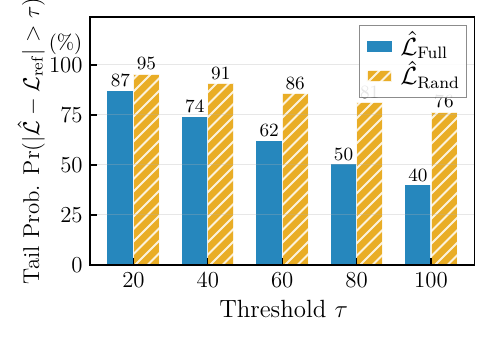}\end{minipage}\hfill
    \begin{minipage}[c]{0.18\linewidth}\centering\includegraphics[width=\linewidth]{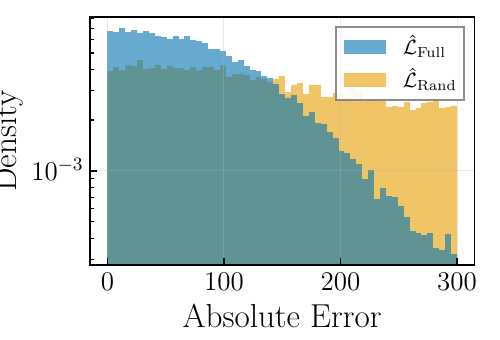}\end{minipage}\hfill
    \begin{minipage}[c]{0.18\linewidth}\centering\includegraphics[width=\linewidth]{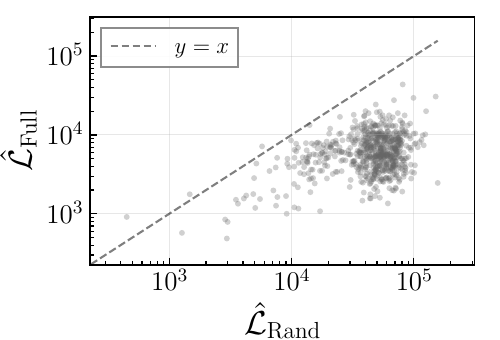}\end{minipage}
    \caption{Detailed estimation quality analysis across five datasets. Rows correspond to datasets (GSM8K, MATH-500, Countdown, Sudoku, and Code); columns correspond to the five analysis types described in the introduction of this section.}
    \label{fig:detailed_all}
\end{figure*}

\subsection{Per-Dataset Details}
\label{subsec:per_dataset_details}

Fig.~\ref{fig:detailed_all} presents all results as a single figure, with five datasets as rows and five types of analysis as columns. We describe the first two columns and the last two columns as grouped pairs below; the Error tail column (the third column) is detailed in Appendix~\ref{subsec:tail_probability}.

\paragraph{Estimation quality (Spearman \texorpdfstring{$\rho$}{} and RMSE).}
The first two columns plot the Spearman rank correlation $\rho$ and RMSE between each estimator and the reference value $\mathcal{L}_{\mathrm{ref}}$ as a function of the Monte Carlo sample count $K$. Both quantities compare the $K$ required for $\hat{\mathcal{L}}_{\text{Full}}$ and $\hat{\mathcal{L}}_{\text{Rand}}$ to achieve comparable estimation quality, with the star marker indicating the result of $\hat{\mathcal{L}}_{\text{Full}}$ at $K = C = 4$. On all five datasets, the $\hat{\mathcal{L}}_{\text{Full}}$ star marker lies at a position where $\hat{\mathcal{L}}_{\text{Rand}}$ requires a much larger $K$ (roughly an order of magnitude larger) to reach comparable Spearman rank correlation and RMSE. This means that the effect observed on GSM8K---``making block-index selection deterministic improves estimation quality''---is consistent across all datasets.

\paragraph{Per-prompt error distribution and variance.}
The last two columns visualize per-prompt error and variance distributions. Error dist.\ (a) shows the density distribution of the estimation error $|\hat{\mathcal{L}} - \mathcal{L}_{\mathrm{ref}}|$ (with the vertical axis on a log scale), comparing the heaviness of the error tail. Error dist.\ (b) plots the per-prompt variances of the absolute error computed from 50 iterations as a scatter plot of $\hat{\mathcal{L}}_{\text{Rand}}$ vs.\ $\hat{\mathcal{L}}_{\text{Full}}$; points lying below the diagonal $y = x$ indicate smaller per-sample variance for $\hat{\mathcal{L}}_{\text{Full}}$. In Error dist.\ (a), the $\hat{\mathcal{L}}_{\text{Full}}$ distribution concentrates toward smaller errors on all five datasets, with the heavy tail suppressed. In Error dist.\ (b), the vast majority of prompts lie below the diagonal $y = x$, showing that the per-sample variance of $\hat{\mathcal{L}}_{\text{Full}}$ is smaller than that of $\hat{\mathcal{L}}_{\text{Rand}}$ on a per-prompt basis. Together, these results indicate that the advantage of our method holds not only at the average level but also at the per-prompt level.

\subsection{Tail Probability of Large Estimation Errors}
\label{subsec:tail_probability}

Whereas Spearman $\rho$ and RMSE summarize the overall trend of the error distribution, the tail probability discussed in this subsection directly compares the frequency of rare large estimation errors. Specifically, for each estimator $\hat{\mathcal{L}} \in \{\hat{\mathcal{L}}_{\text{Full}}, \hat{\mathcal{L}}_{\text{Rand}}\}$, we evaluate the tail probability $\Pr(|\hat{\mathcal{L}} - \mathcal{L}_{\mathrm{ref}}| > \tau)$ at each threshold $\tau$. Measurements are computed with $K = 4$ from 50 independent iterations of each prompt, using the absolute error against the reference value. The third column (Error tail) of Fig.~\ref{fig:detailed_all} shows the results on all five datasets. For example, on GSM8K at $\tau = 100$, $\hat{\mathcal{L}}_{\text{Rand}}$ exceeds the threshold with probability $45.9\%$, while $\hat{\mathcal{L}}_{\text{Full}}$ does so with only $4.4\%$; the ratio between the two grows from $1.4\times$ at $\tau = 20$ to $10.5\times$ at $\tau = 100$ as the threshold increases. This shows that the error distribution of $\hat{\mathcal{L}}_{\text{Rand}}$ has a heavier tail, and the same trend holds on the other datasets.

\section{Variance Reduction at the Policy-Gradient Level}
\label{app:gradient_variance_argument}

This section examines, through both a theoretical argument and empirical measurements, how far the variance reduction demonstrated at the estimator level carries over to the between-group variance of the policy gradient.
Here, the between-group variance refers to the variance arising from the stochastic selection of the block index $b \in \{0, \ldots, C-1\}$, in analogy with the loss-level definition $\sigma_{\mathrm{between}}^2$ in (\ref{eq:sig_between}).

\subsection{From the Estimator to the Policy Gradient}
\label{subapp:gradient_theory}
The variance reduction shown so far in Section~\ref{sec:variance_reduction} and Section~\ref{sec:analysis} is an effect on the estimator $\hat{\mathcal{L}}$ itself.
Training stability depends on the variance of the policy gradient $\nabla_\theta \mathcal{J}_{\text{Block}}$, so we examine the relationship between the two.
By making block selection deterministic, our method eliminates the between-group variance from both the loss and the gradient.
The GRPO advantage $A^j$ differs across responses $j$, but the same $A^j$ multiplies every block sample within a single response, so it does not affect the between-group variance ratio.
In our experiments $\beta = 0$ (Appendix~\ref{app:training_details}, Table~\ref{tab:common_hyperparams}), so the KL regularization term also does not contribute.
Therefore, the variance reduction shown at the estimator level is preserved for the between-group variance of the policy gradient as well, in the sense that the variance ratio is at most $1$.

\subsection{Experimental Setup of the Empirical Measurement}
\label{subapp:measurement_protocol}
The argument above is a theoretical prediction; whether the sample variance of the gradient is actually reduced in line with this prediction during real training is an empirical question that must be examined independently.
Table~\ref{tab:fa_grad_variance}, reported in this section, and Table~\ref{tab:fa_variance_decomp}, reported in Appendix~\ref{app:variance_ratio_post}, are two analyses that extract different quantities from a common measurement procedure.
Specifically, for each training step in $\{1{,}000, 2{,}000, 3{,}000, 4{,}000\}$ of the $\hat{\mathcal{L}}_{\text{Full}}$ run in the main experiment (LLaDA-8B-Instruct, $C = K = 4$), we sample $N = 100$ prompts from the GSM8K and MATH-500 evaluation sets at each checkpoint.
For each prompt, we generate $g = 4$ responses from the policy (temperature $= 0.9$, identical to the GRPO sampling of the main experiment), compute rewards using the rule-based reward function, and obtain the unnormalized GRPO advantage $A^j = R^j - \bar{R}$.
In evaluation mode (dropout off), for each prompt we independently sample $\hat{\mathcal{L}}_{\text{Full}}$ and $\hat{\mathcal{L}}_{\text{Rand}}$ $S = 50$ times, and denote the corresponding policy-gradient samples by $\hat{\mathcal{G}}_{\text{Full}}$ and $\hat{\mathcal{G}}_{\text{Rand}}$ ($\nabla_\theta \mathcal{J}_{\text{Block}}$ computed using the Full and Rand estimators, respectively). We also record the per-block log-likelihood $\ell_{b, t_b, \mathcal{M}_{t_b}}$.
Prompts for which all $g$ responses receive the same reward yield $A^j = 0\;\forall j$ and a zero policy gradient; the values reported in Table~\ref{tab:fa_grad_variance} and Table~\ref{tab:fa_variance_decomp} therefore aggregate over the remaining prompts after excluding these.

\subsection{Empirical Variance Ratio of the Policy Gradient}
\label{subapp:gradient_empirical}
Using the $S = 50$ gradient samples $\hat{\mathcal{G}}_{\text{Full}}$ and $\hat{\mathcal{G}}_{\text{Rand}}$ from the experimental setup, we compute the per-prompt gradient variance and report the ratio $\mathrm{Var}[\hat{\mathcal{G}}_{\text{Full}}] / \mathrm{Var}[\hat{\mathcal{G}}_{\text{Rand}}]$ averaged over the remaining prompts (after exclusion) in Table~\ref{tab:fa_grad_variance}.
As shown in Table~\ref{tab:fa_grad_variance}, the ratio is below $1$ at every measured checkpoint.
However, the ratio ranges from 0.79 to 0.96, which is closer to 1 than the estimator-level ratio of 0.10 to 0.16 measured at the same checkpoints (Table~\ref{tab:fa_variance_decomp}).

\begin{table}[t]
    \small
    \centering
    \setlength{\tabcolsep}{4pt}
    \begin{tabular}{l cccc}
        \toprule
        & \multicolumn{4}{c}{Training step} \\
        \cmidrule(lr){2-5}
        Task & 1{,}000 & 2{,}000 & 3{,}000 & 4{,}000 \\
        \midrule
        GSM8K & 0.811 & 0.915 & 0.889 & 0.961 \\
        MATH-500 & 0.791 & 0.926 & 0.825 & 0.915 \\
        \bottomrule
    \end{tabular}
    \caption{Sample variance ratio of the policy gradient $\mathrm{Var}[\hat{\mathcal{G}}_{\text{Full}}] / \mathrm{Var}[\hat{\mathcal{G}}_{\text{Rand}}]$. Computed for the four training checkpoints and the two tasks of the main experiment ($C = K = 4$), aggregating over the remaining prompts after excluding those with $A^j = 0\;\forall j$ from the $N = 100$ samples. Each prompt contributes $S = 50$ gradient samples. A ratio below $1$ indicates that the variance of Full is smaller than that of Rand.}
    \label{tab:fa_grad_variance}
\end{table}

\section{Estimator Variance Ratio at Mid-Training Checkpoints}
\label{app:variance_ratio_post}

Table~\ref{tab:variance_ratio} in Section~\ref{subsec:variance_ratio} reports the estimator variance ratio $\mathrm{Var}[\hat{\mathcal{L}}_{\text{Full}}] / \mathrm{Var}[\hat{\mathcal{L}}_{\text{Rand}}]$ in (\ref{eq:var_ratio}) on the base model.
In this section, we compute the same quantity for four mid-training checkpoints $\times$ two tasks.
The measurement setup is shared with Appendix~\ref{subapp:measurement_protocol}: using the $S = 50$ samples of $\hat{\mathcal{L}}_{\text{Full}}$, we estimate the per-block variance and mean via the same procedure as in Section~\ref{subsec:variance_ratio}, and report the variance ratio $1 - \lambda = \bar{\sigma}_{\text{within}}^2 / (\sigma_{\text{between}}^2 + \bar{\sigma}_{\text{within}}^2)$ at $K = C = 4$ in Table~\ref{tab:fa_variance_decomp}.
Directly comparing Table~\ref{tab:variance_ratio} (base model) with Table~\ref{tab:fa_variance_decomp} (mid-training checkpoints) lets us check whether the variance-ratio expression takes a comparable value at mid-training checkpoints.

\begin{table}[t]
    \small
    \centering
    \setlength{\tabcolsep}{4pt}
    \begin{tabular}{l cccc}
        \toprule
        & \multicolumn{4}{c}{Training step} \\
        \cmidrule(lr){2-5}
        Task & 1{,}000 & 2{,}000 & 3{,}000 & 4{,}000 \\
        \midrule
        GSM8K & 0.115 & 0.113 & 0.138 & 0.134 \\
        MATH-500 & 0.095 & 0.155 & 0.104 & 0.132 \\
        \bottomrule
    \end{tabular}
    \caption{Estimator variance ratio $1 - \lambda = \bar{\sigma}_{\text{within}}^2 / (\sigma_{\text{between}}^2 + \bar{\sigma}_{\text{within}}^2)$ ($K = C = 4$) at the mid-training checkpoints of the main experiment. Computed from the variance decomposition (\ref{eq:var_ratio}) of the per-block log-likelihood $\ell_{b, t_b, \mathcal{M}_{t_b}}$ and aggregated over the prompts excluding those with $A^j = 0\;\forall j$. Directly comparable to Table~\ref{tab:variance_ratio} on the base model. Smaller values indicate larger variance reduction.}
    \label{tab:fa_variance_decomp}
\end{table}

As shown in Table~\ref{tab:fa_variance_decomp}, the variance ratio lies in the range $0.10$--$0.16$ across all checkpoints and both tasks; it is close to the values $0.08$--$0.12$ on the base model in Table~\ref{tab:variance_ratio} and well below $1$, so the variance-reduction effect of $\hat{\mathcal{L}}_{\text{Full}}$ is preserved at every measured checkpoint during training.

\section{Relation to Classical Sampling Techniques}
\label{app:related_extension}

This section elaborates on the relation to classical sampling techniques mentioned in Section~\ref{sec:related_work}.

\paragraph{Relation to stratified sampling.}
Stratified sampling is a classical variance-reduction technique that partitions the population into non-overlapping strata and draws samples independently from each stratum.
Our $\hat{\mathcal{L}}_{\text{Full}}$ at $K = C$ is equivalent to stratified sampling with one sample drawn per stratum.
The contribution of this work is to introduce the perspective of Monte Carlo estimator design into the blockwise generation structure of MDLM.

\paragraph{Extension to importance sampling.}
A natural extension of stratified sampling is importance sampling based on block variance.
The idea is to concentrate weight on high-variance blocks to lower the overall estimation variance, but in the context of policy optimization a different issue can arise.
When the weights concentrate on a small number of blocks, the policy is barely updated on the other blocks; learning becomes imbalanced, and the policy adapts only to those specific blocks, which risks policy collapse (a phenomenon in which over-adaptation to specific states degrades policy performance).
This paper adopts uniform full coverage as a more stable choice, and stabilizing weighted-sampling strategies remains future work.

\end{document}